\documentclass[twocolumn]{svjour3}
\usepackage{epsfig}
\usepackage{graphicx}
\usepackage{amsmath}
\usepackage{amssymb}
\usepackage{bbm}
\usepackage{diagbox}
\usepackage{tabularx}
\usepackage{xcolor}
\usepackage[ruled]{algorithm2e}
\usepackage{colortbl}
\usepackage{multirow}
\usepackage{bbding}
\usepackage{subfig}
\usepackage{ragged2e}
\usepackage{colortbl}
\usepackage{arydshln}
\usepackage{enumerate}
\usepackage{pifont}
\usepackage{makecell}
\usepackage{enumitem}
\usepackage{natbib}

\usepackage{hyperref}
\hypersetup{hidelinks,
	colorlinks=true,
	allcolors=black,
	pdfstartview=Fit,
	breaklinks=true}

\usepackage{xspace}
\makeatletter
\DeclareRobustCommand\onedot{\futurelet\@let@token\@onedot}
\def\@onedot{\ifx\@let@token.\else.\null\fi\xspace}

\def\eg{\emph{e.g}\onedot} 
\def\ie{\emph{i.e}\onedot} 
 
\def\etc{\emph{etc}\onedot} 
\newcommand{\blue}[1]{\textcolor{black}{#1}}

\definecolor{mygray}{gray}{.9}

\def\etal{\emph{et al}\onedot}
\def\eg{\emph{e}\onedot \emph{g}\onedot}
\newcolumntype{P}[1]{>{\centering\arraybackslash}p{#1}}
\begin{document}

\title{Transformer for Object Re-Identification: A Survey}


\author{Mang Ye, Shuoyi Chen, Chenyue Li, Wei-Shi Zheng,\\ David Crandall, Bo Du}

\institute{
M. Ye, SY. Chen, CY. Li and B. Du are with the National Engineering Research Center for Multimedia Software, School of Computer Science, Hubei Luojia Laboratory, Wuhan University, Wuhan, China. \\
WS. Zheng is with the School of Data and Computer Science, Sun Yat-sen University, Guangzhou, China. \\
D. Crandall is with the Luddy School of Informatics, Computing, and Engineering, Indiana University.
}

\maketitle

\begin{sloppypar}
\begin{abstract}
Object Re-identification (Re-ID) aims to identify specific objects across different times and scenes, which is a widely researched task in computer vision.
For a prolonged period, this field has been predominantly driven by deep learning technology based on convolutional neural networks. In recent years, the emergence of Vision Transformers has spurred a growing number of studies delving deeper into Transformer-based Re-ID, continuously breaking performance records and witnessing significant progress in the Re-ID field. 
Offering a powerful, flexible, and unified solution, Transformers cater to a wide array of Re-ID tasks with unparalleled efficacy. 
This paper provides a comprehensive review and in-depth analysis of the Transformer-based Re-ID. In categorizing existing works into Image/Video-Based Re-ID, Re-ID with limited data/annotations, Cross-Modal Re-ID, and Special Re-ID Scenarios, we thoroughly elucidate the advantages demonstrated by the Transformer in addressing a multitude of challenges across these domains. Considering the trending unsupervised Re-ID, we propose a new Transformer baseline, UntransReID, achieving state-of-the-art performance on both single/cross modal tasks. 
For the under-explored animal Re-ID, we devise a standardized experimental benchmark and conduct extensive experiments to explore the applicability of Transformer for this task and facilitate future research. Finally, we discuss some important yet under-investigated open issues in the large foundation model era, we believe it will serve as a new handbook for researchers in this field. A periodically updated website will be available at \url{https://github.com/mangye16/ReID-Survey}.
\end{abstract}

\begin{figure*}[t]
    \centering    \includegraphics[width=0.46\linewidth]{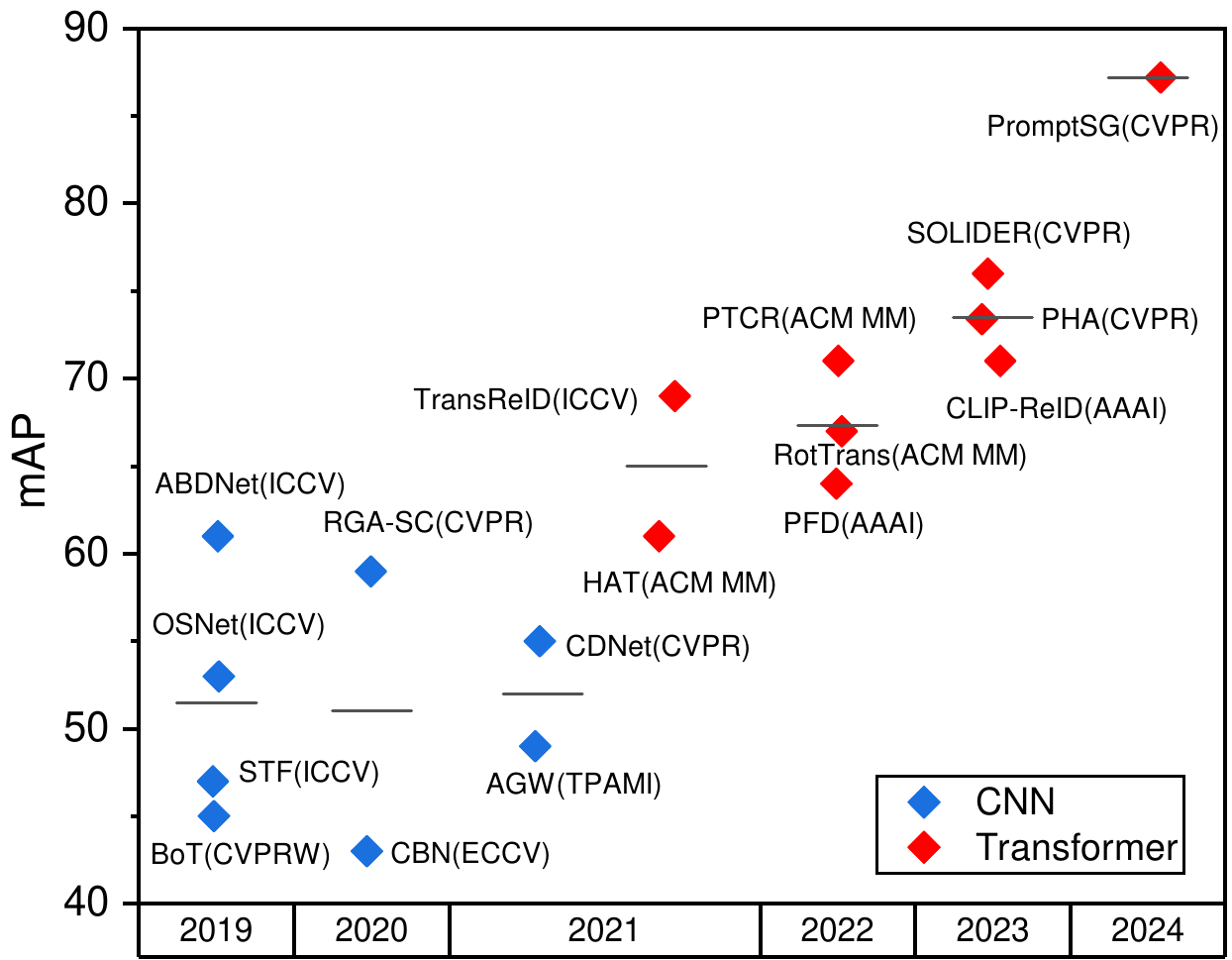} 
    \includegraphics[width=0.46\linewidth]{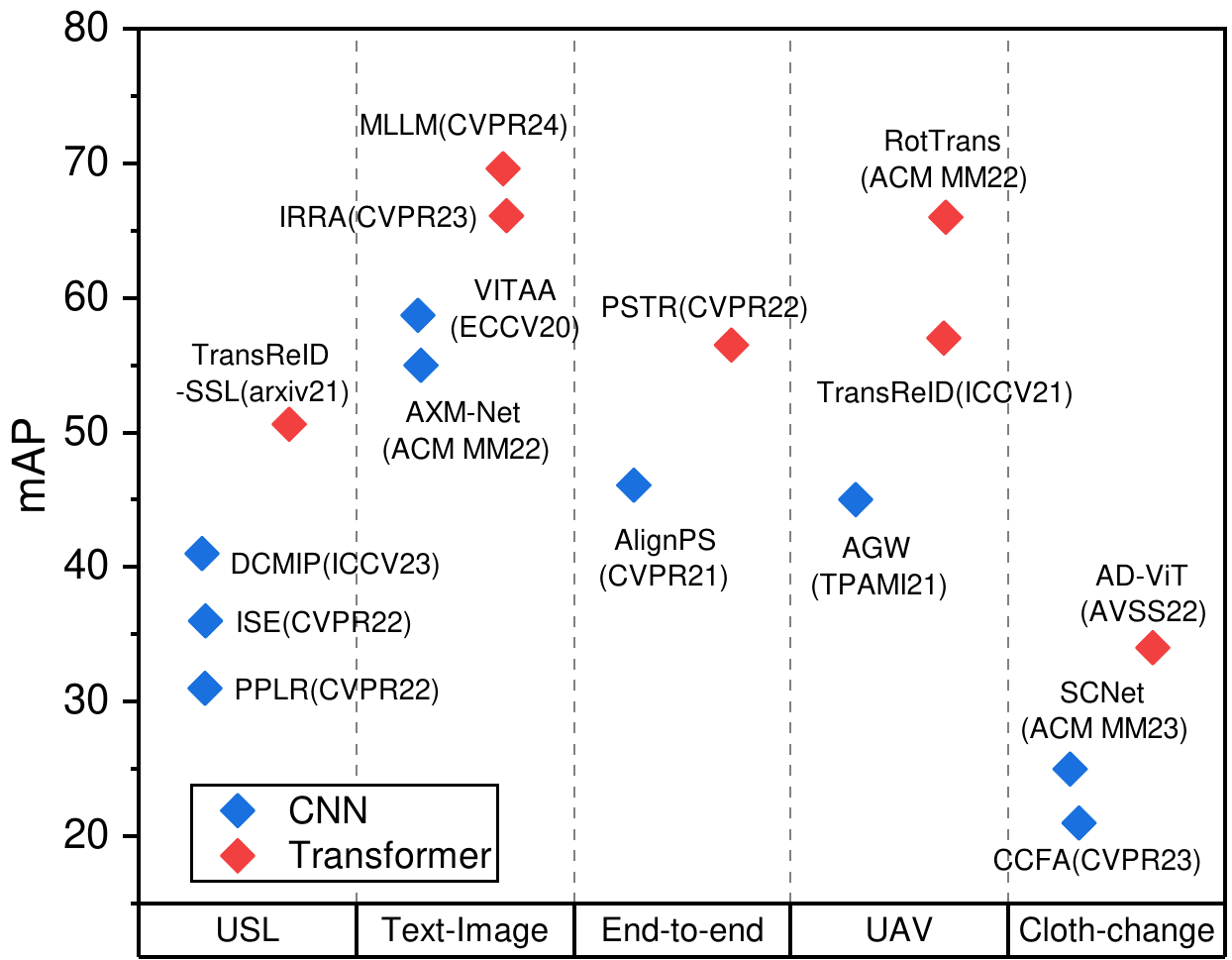}
    \caption{(1) We show the performance of recent state-of-the-art methods on the widely-used person Re-ID dataset MSMT17 (left). The transformer-based methods have achieved a comprehensive lead in accuracy since 2021, while the CNN-based method for single-modal image Re-ID has not been studied. (2) We show state-of-the-art results of representative works in different Re-ID tasks: unsupervised (USL) Re-ID on MSMT17 \citep{wei2018person} dataset, Text-Image on CUHK-PEDES \citep{li2017person}, end-to-end person search on PRW \citep{zheng2017person}, Re-ID in UAVs on PRAI-1581 \citep{zhang2020person}, and cloth-changing Re-ID on LTCC \citep{qian2020long}.}
    \vspace{-2mm}
    \label{fig:intro}
\end{figure*}

\section{Introduction}
\label{sec:introduction}
The object re-identification (Re-ID), which aims at matching the same object (person~\citep{gray2007evaluating}, vehicles~\citep{sun2004vehicle}, etc) across multiple different views\citep{arxiv17triplet,zhong2017re}. Over the years, object Re-ID has attracted considerable attention as a significant research area \citep{ye2021deep,zheng2015scalable,ahmed2015improved}, 
expanding the application scope of tasks such as object detection, tracking, and recognition. It holds substantial practical application value in domains such as intelligent surveillance, smart cities, and the preservation of natural ecosystems.
In recent years, research in the Re-ID field, particularly concerning subjects like persons and vehicles, has undergone profound development and has achieved notable success in conventional settings. Moreover, Re-ID encompasses a diverse array of object categories, including animals, buildings, and more. In order to better address real-world application demands, existing Re-ID research is gradually shifting its focus towards open-world scenarios. This entails tackling challenges such as managing large-scale data with limited annotations \citep{xuan2021intra,zhang2022implicit,yu2019unsupervised}, diverse data modalities \citep{ye2021channel, wu2023unsupervised}, generalization of unknown scenarios \citep{jin2020style, zhao2021learning},  as well as tackling specialized applications like long-term Re-ID or group Re-ID \citep{chen2022rotation, rao2021self, xiao2018group}. These emerging research directions hold rich potential for further advancing Re-ID and facilitating its practical deployment.

Benefiting from the development of deep learning \citep{cheng2023discriminative, cheng2024progressive}, the studies in the Re-ID field have been dominated by deep Convolutional Neural Networks (CNNs) for a long time \citep{ye2021deep,zheng2016person}. Nevertheless, the emergence of the Vision Transformer \citep{vaswani2017attention, dosovitskiyimage} has changed this situation. Transformer is a network architecture dispensing with recurrence and convolutions that relies entirely on attention mechanisms to model the global dependencies between inputs and outputs. Transformer has been introduced into the field of computer vision as a breakthrough, demonstrating remarkable performance in various visual tasks, including Re-ID. Intuitively, some works try to directly replace CNNs with existing vision transformers \citep{dosovitskiyimage, liu2021swin, wang2021pyramid} as the feature extractor, which significantly improves the accuracy of Re-ID \citep{he2021transreid, chen2022rotation, cao2022pstr, 
zhang2022interlaced, comandur2022sports}. 
Unlike CNNs, Vision Transformer (ViT) \citep{dosovitskiyimage} architecture imposes little structural bias to guide representation learning, which allows for diverse learning strategy design and broad task applicability \citep{walmer2023teaching}. This allows the Transformer to be flexibly applied to various scenarios including unsupervised Re-ID, multimodal Re-ID, etc. Furthermore, ViT treats an image as a sequence of patch tokens rather than processing images pixel-by-pixel, which allows inputs of various sizes to be accepted without additional adjustments. The tokenization feature exhibits strong compatibility for personalized design and offers flexibility in organizing information \citep{xu2023multimodal, han2022survey, naseer2021intriguing}. These advantages facilitate the seamless integration of Transformers with Re-ID-specific designs, and novel research ideas for Re-ID that leverage the unique properties of transformers continue to emerge \citep{jiang2023cross, luo2021self, rao2023transg, li2022pyramidal, chen2022rotation}. In recent years, Transformer-based Re-ID research has consistently set new records in recognition accuracy, showing a trend of being significantly superior to CNN-based methods in many aspects, as shown in Fig. \ref{fig:intro}. Due to the rapid proliferation of transformer-based Re-ID models, it is becoming progressively challenging to stay abreast of the latest advancements. Consequently, there is an urgent need for a comprehensive survey of existing Transformer-based works, which would greatly benefit the community in the new era.

\begin{figure*}[!t]
    \centering
    \includegraphics[width=\linewidth]{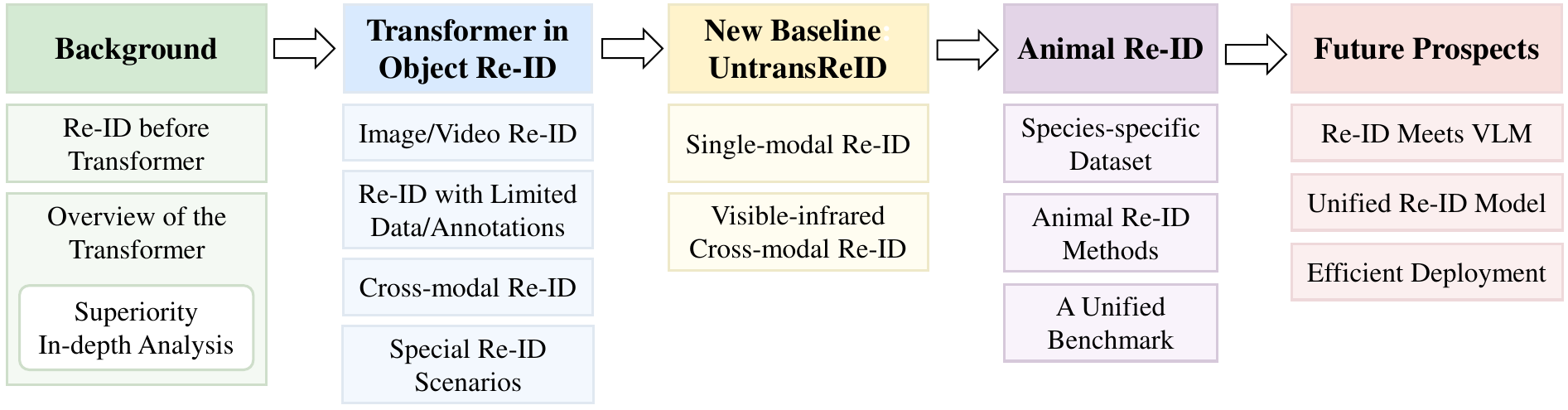}
    \caption{\blue{An overview of the framework structure for the survey, illustrating key sections and their interrelationships. }}
    
    \label{fig:survey_structure}
    \vspace{-3mm}
\end{figure*}

Existing Re-ID surveys \citep{ye2021deep, zheng2016person, khan2019survey, wang2019beyond} predominantly focus on deep learning methods based on CNNs and tend to narrow their scope to specific objects, with a primary emphasis on persons or vehicles. 
On the contrary, this survey is mainly oriented towards the application of emerging transformer technology in Re-ID and covers a wider range of objects (persons, vehicles, and animals), which is more innovative and comprehensive. Recognizing the significant potential and promise demonstrated by numerous Transformer-based studies in various vision applications, we systematically organize and review the growing research works on Transformer for Re-ID in recent years to gain valuable insights. Differing from existing surveys, the primary contributions of our survey are as follows: 
\begin{itemize}
\item We conduct an in-depth analysis of the strengths of Transformer and summarize the research efforts since its introduction into the Re-ID field across four extensively studied Re-ID directions, including image/video-based Re-ID, Re-ID with limited data/annotations, cross-modal Re-ID, and special Re-ID scenarios. It demonstrates the success of Transformer in Re-ID and underscores its potential for future advancements.
\item We introduce a Transformer-based unsupervised baseline for trending unsupervised Re-ID, leveraging the Transformer architecture, which remains relatively underexplored in existing works. Our proposed method demonstrates competitive performance across both single- and cross-modal unsupervised Re-ID tasks.
\item We particularly delve into animal re-identification, an area that has received significantly less attention compared to persons and vehicles. It presents numerous challenges and unresolved issues. We develop unified experimental standards for animal Re-ID and evaluate the feasibility of employing Transformer in this context, laying a solid foundation for future research.
\end{itemize}

The rest of this survey is organized as follows: In \S \ref{sec:background}, we briefly review the development of the Re-ID field before the Transformer era while introducing the Transformer in vision and providing a detailed analysis of its numerous advantages. \blue{ A comprehensive analysis of Transformer in Re-ID is presented in \S \ref{sec:Transformer Re-ID}. A powerful Transformer baseline for single/cross-modal unsupervised Re-ID is proposed in \S \ref{sec:new_baseline}.} The progress in Animal Re-ID and the evaluation of the applicability of Transformer to this task are introduced in \S \ref{sec:animal_Re-ID}. \blue{In Fig. \ref{fig:survey_structure}, we present the overall framework structure of this survey, outlining the key sections and their interconnections.}


\section{Background}
\label{sec:background}


\begin{figure*}[!t]
    \centering
    \includegraphics[width=\linewidth]{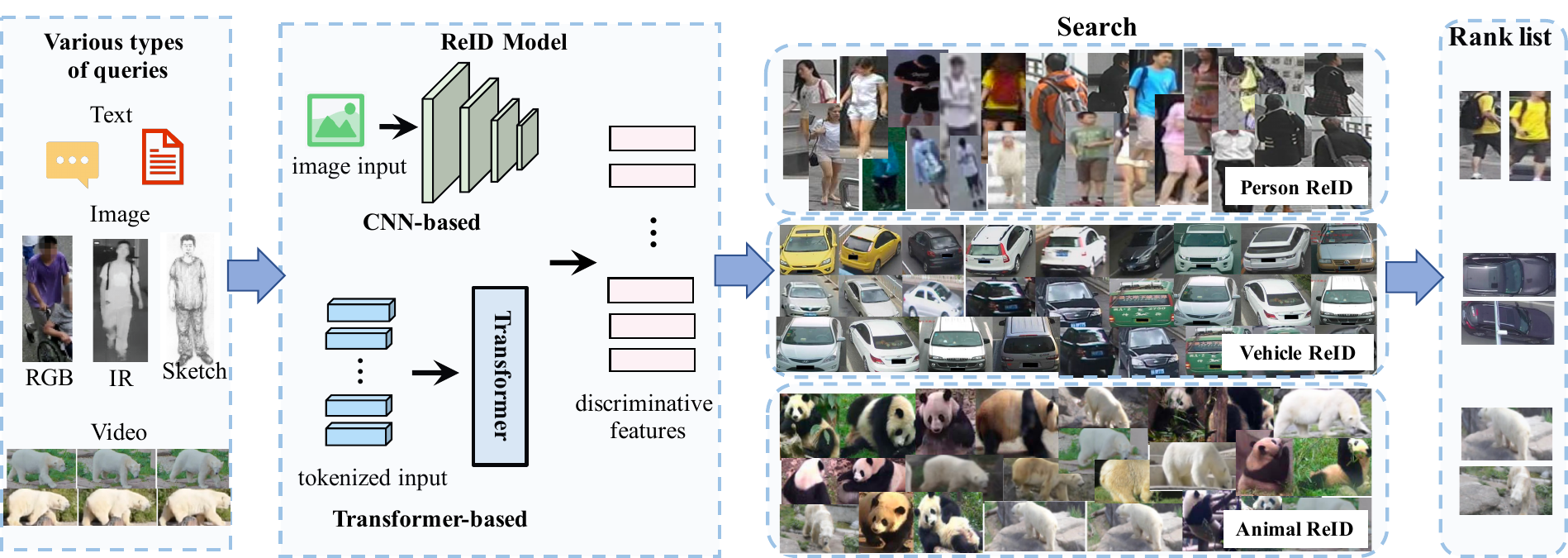}
    \caption{\textbf{General object Re-ID process.} Given a query that can be any type of image, text, video, \etc, the goal of Re-ID is to search for the specific object from gallery data collected by different cameras. }
    
    \label{fig:process}
    \vspace{-3mm}
\end{figure*}

\subsection{A Brief Review of Re-ID Before Transformers}
This subsection begins by summarizing the fundamental definition and challenges of object Re-ID, followed by an introduction to commonly used datasets and evaluation metrics (\S \ref{sec:Re-ID}). Then, we give a general review of previous CNN-dominated Re-ID research works and discuss the limitations of CNN-based Re-ID methods in some aspects (\S \ref{sec:Re-ID_methods}).
\label{sec:Re-ID_background}
\subsubsection{Object Re-Identification}
\label{sec:Re-ID}

\textbf{Definition.} Given a query $q$, the goal of object re-identification is to retrieve specific objects from a gallery set $\mathcal{G} =\{g_i | i=1,2,\cdots, N\}$ of $N$ descriptors. An important feature of Re-ID is to identify the objects in different cameras with non-overlapping views. The query $q$ can be an image, a video sequence, a text description, a sketch, a combination of different forms, \etc \citep{liu2016deep, zheng2016person, liu2016large, chen2017person, ye2021channel, jiang2023cross, chen2023towards}. The identity of the query $q$ can be formulated as:
\begin{equation}
    I = \mathop{\arg\min}_{g_i}dis(q,g_i),g_i \in \mathcal{G},
\end{equation}
where $dis(\cdot , \cdot)$ is an arbitrary distance metric.

\textbf{Challenges.} The flow of a basic Re-ID system is shown in Fig. \ref{fig:process}. \blue{To provide a detailed overview of the challenges in Re-ID tasks, we have divided the discussion into two main categories. (1) \textit{Object Types.} The objects that are widely studied currently include persons and vehicles. Person Re-ID involves the challenge of individuals frequently changing clothes, accessories, and appearance. This variability can result in significant differences in visual appearance, making it difficult for Re-ID models to consistently identify the same individual over time. Therefore, models are required to disregard changes in clothing and focus on more invariant features, such as body shape or gait. Additionally, human body posture can vary considerably due to movement and camera angles. Discriminative regions, such as the face, are often difficult to capture under these circumstances. In vehicle Re-ID, the primary challenge lies in the significant intra-class similarity. Vehicles of the same brand, model, and color often appear nearly identical. Even minor differences, such as subtle scratches or decals, may be crucial for distinguishing one vehicle from another, but these differences can be difficult to detect or may not always be visible. Additionally, vehicles can look different when viewed from various angles (e.g., front, rear, or side). Unlike persons whose body shape remains relatively constant, the geometry and key identifying features of a vehicle can vary substantially depending on the viewpoint, making it challenging for Re-ID systems to generalize across multiple perspectives.  (2) \textit{Application Scenarios.} The early Re-ID tasks mainly focused on the pre-detected object bounding boxes obtained from the original image or video and used the appearance information to match the corresponding individual. Due to the complex conditions of image acquisition, the difficulty of Re-ID in this period involves occlusion, illumination variation, resolution difference, camera view variation, and background clutter \citep{zheng2016person,khan2019survey,ahmed2015improved}. Furthermore, new challenges continue to emerge with the urgent growth of practical application requirements \citep{wu2023unsupervised, yang2022augmented, yu2019unsupervised}. For example, with the widespread application of drone surveillance recently \citep{zhang2020person, wang2019vehicle, teng2021viewpoint}, discriminative information has been greatly reduced under extreme bird's-eye view angles \citep{li2021uav, kumar2020p}. In the data processing stage, for special cases where conventional visible light images are not available, valid object information might be represented by different modalities such as infrared images, text descriptions, sketches, and depth images \citep{chen2023towards, ye2023channel, zhai2022triReID}. The cross-modal gaps of object Re-ID in the modal heterogeneous scene lead to great intra-class differences across varying modalities \citep{li2017person, wang2020vitaa}. In addition, due to the artificial cross-camera correlation of objects, high labeling costs \citep{ li2019unsupervised, fu2021unsupervised, cheng2022hybrid} and unavoidable noise problems \citep{ye2021collaborative} make it difficult to achieve large-scale expansion \citep{ye2020augmentation, lin2019bottom, cho2022part}. In the retrieval phase, the varying real environmental domains may cause the inapplicability of the Re-ID model \citep{bai2021person30k,ni2022meta}, and long-term Re-ID suffers from appearance information changes \citep{qian2020long, fan2020learning}. 
}

\blue{\textbf{Datasets.} In Table \ref{all_datasets}, we provide a comprehensive summary of widely utilized datasets for various Re-ID tasks. These datasets encompass a range of challenges specific to different object types, such as persons and vehicles, offering diverse conditions for evaluating Re-ID algorithms. By aggregating key details, including the number of identities, images, and cameras, this table serves as a valuable reference for understanding the scale, diversity, and complexity of datasets frequently employed in Re-ID research.}

\blue{\textbf{Evaluation Metrics.}
In the evaluation of Re-ID models, two commonly adopted metrics are the Cumulative Matching Characteristic (CMC) and the mean Average Precision (mAP).
The CMC curve measures the probability that a correct match for a given query appears within the top-K ranked results. It is particularly effective in ranking-based evaluation scenarios, providing insights into how well a model retrieves the correct identity from a gallery. The CMC at rank-1 is often emphasized, as it reflects the model’s ability to correctly identify the target in the top position, making it a critical metric for real-world Re-ID applications where quick and accurate identification is crucial.
On the other hand, mAP provides a more comprehensive evaluation by considering both precision and recall over the entire ranked list. It computes the average precision for each query and then calculates the mean across all queries. Unlike CMC, mAP is sensitive to both the ranking and the completeness of retrieved results, making it a robust metric for cases where the correct identity might appear lower in the ranking list. This metric is particularly valuable in scenarios where it is important not only to rank the correct match highly but also to retrieve all relevant matches with high precision.
Together, CMC and mAP provide a well-rounded assessment of Re-ID models, reflecting their performance in ranking accuracy and retrieval quality. }

\blue{Besides, mINP (mean Inverse Negative Penalty) \citep{ye2021deep} is a newly proposed metric designed to enhance the evaluation of Re-ID systems by focusing on the rank position of the hardest correct match, which is crucial for effective tracking in multi-camera networks. Unlike traditional metrics like CMC and mAP, which may not accurately reflect the challenges of identifying all correct matches, mINP quantifies the penalty incurred when searching for the hardest match. This metric is computationally efficient and can be easily integrated into existing CMC/mAP evaluation frameworks. While it may exhibit smaller value differences with larger gallery sizes, mINP still effectively indicates the relative performance of a Re-ID model, serving as a valuable complement to conventional metrics.}

\begin{table*}
\footnotesize
\centering
\renewcommand{\arraystretch}{1.2}
\caption{\blue{Summary of Commonly Used Datasets for Diverse Re-ID Tasks.}}
\label{all_datasets}
\resizebox{\textwidth}{!}{%
\begin{tabular}{c | c c c c c}
\hline
Dataset & Year & Images/Boxes  & Identities & Object Type &  Characterization  \\
\hline
\multicolumn{6}{c}{\textbf{Image/Video Based Re-ID}} \\ 
\hline
MSMT17\citep{wei2018person} & 2018 & 126,441 & 4,101 & Person & Image  \\
DukeMTMC-ReID\citep{zheng2017unlabeled} & 2017 & 36,441 & 1,812 & Person & Image \\
Market1501\citep{zheng2015scalable} & 2015 & 32,217 & 1,501 & Person & Image \\
CUHK03\citep{li2014deepreid} & 2014 & 13,164 & 1,467 & Person & Image \\
MARS\citep{zheng2016mars} & 2016 & 1,191,003 & 1,261 & Person & Video \\
LPW\citep{song2018region} & 2018 & 592,438 & 2,731 & Person & Video  \\

CityFlow\citep{tang2019cityflow} & 2019 & 56,277 & 666 & Vehicle & Image  \\
VERI-Wild\citep{lou2019veri} & 2019 & 416,314 & 40,671 & Vehicle & Image \\
PKU-VD(VD1/VD2)\citep{yan2017exploiting} & 2017 & 846,358/807,260 & 1,232/1,112 & Vehicle & Image \\
VehicleID\citep{zapletal2016vehicle} & 2016 & 221,763 & 26,267 & Vehicle & Image  \\
VeRi-776\citep{liu2016deep} & 2016 & 49,357 & 776 & Vehicle & Image \\

\hline
\multicolumn{6}{c}{\textbf{Cross-modality Re-ID}} \\ 
\hline
CUHK-PEDES\citep{li2017person} & 2017 & 40,206 (image), 80,422 (text) & 13,003 & Person & Image-text  \\
ICFG-PEDES\citep{ding2021semantically} & 2021 & 54,522 (image), 54,522 (text) & 4,102 & Person & Image-text  \\
RSTPReid\citep{zhu2021dssl} & 2021 & 20,505 (image), 41,010 (text) & 4,101 & Person & Image-text  \\
RegDB \citep{nguyen2017person} & 2017 & 4,120 (RGB), 4,120 (thermal) & 412 & Person & Visible-thermal   \\
SYSU-MM01\citep{wu2017rgb} & 2017 & 20,284 (RGB), 9,929 (infrared) & 296 & Person & Visible-infrared \\
PKU SketchRe-ID\citep{pang2018cross} & 2018 & 400 & 200 & Person & Sketch \\

\hline
\multicolumn{6}{c}{\textbf{ReID with Limited Data/Annotations}} \\ 
\hline
LUPerson\citep{fu2021unsupervised} & 2021 & 4M & 200K & Person & Unlabelled Image \\
VehicleX\citep{yao2020simulating} & 2020 & $\infty$ & 1,362 & Vehicle & Synthetic data \\
PersonX\citep{sun2019dissecting} & 2019 & 45,576 & 1,266 & Person & Synthetic data \\
UnrealPerson\citep{zhang2021unrealperson} & 2021 & 120,000 & 3,000 & Person & Synthetic data \\
WePerson\citep{li2021weperson} & 2021 & 4,000,000 & 1,500 & Person & Synthetic data \\
\hline
\multicolumn{6}{c}{\textbf{Special Re-ID Scenarios}} \\ 
\hline

UAV-Human  \citep{li2021uav} & 2021 & 41,290 & 1,144 & Person  & UAV \\
PRAI-1581 \citep{zhang2020person} & 2020 & 39,461 & 1,581 & Person  & UAV \\
VRAI\citep{wang2019vehicle} & 2019 & 137,613 & 13,022 & Vehicle & UAV \\
UAV-VeID \citep{teng2021viewpoint} & 2021 & 41,917 & 4,601 & Vehicle & UAV \\

Partial-REID\citep{he2021partial} & 2021 & 600 & 60 & Person  & Occluded \\
Occluded-DukeMTMC\citep{miao2019pose} & 2019 & 35,489 & 2,331 & Person  & Occluded \\

Occluded-REID \citep{zhuo2018occluded} & 2018 & 2,000 & 200 & Person  & Occluded \\
DeepChange\citep{xu2023deepchange} & 2023 & 178K & 1,121 & Person & Cloth-changing  \\
LTCC\citep{qian2020long} & 2020 & 17,119 & 152 & Person & Cloth-changing  \\
PRCC\citep{yang2019person} & 2019 & 33,698 & 221 & Person & Cloth-changing  \\

CUHK-SYSU\citep{xiao2017joint} & 2017 & 18,184 &  8,432 & Person & Person Search  \\
PRW\citep{zheng2017person} & 2017 & 5,704 & 482 & Person & Person Search \\
CSG\citep{yan2020learning} & 2020 & 3,839 &  1,558(group classes) & Person & Group Re-lD \\
DukeMTMC Group\citep{lin2019group} & 2019 & 354 &  177(group classes) & Person & Group Re-lD\\
RoadGroup\citep{lin2019group} & 2019 & 324 &  162(group classes) & Person & Group Re-lD \\
\hline
\end{tabular}%
}
\end{table*}

\subsubsection{CNN-based Re-ID Methods}
\label{sec:Re-ID_methods}
Under the mainstream trend of deep learning, the object Re-ID steps are generalized as different steps, including data processing, model training, and descriptor matching. Most existing methods take training a strong Re-ID model as the core goal. In fact, CNNs have dominated Re-ID studies for a long period. In this section, we focus on reviewing the progress of object Re-ID which is highly related to CNNs. Considering different application requirements, Ye \textit{et al.} proposed to divide Re-ID technology into two subsets, closed-world and open-world \citep{ye2021deep}. 

Closed-world refers to supervised learning methods based on well-labeled visible images captured by common video surveillance \citep{liu2016large}. With the aid of labels, many approaches model Re-ID as a classification, verification, or metric learning problem, using CNNs (\ie, ResNet\citep{he2016deep}) to learn discriminative feature representations from the training data \citep{zheng2017discriminatively, luo2019strong, liu2016deep}. On the basis, learning local features such as image slices \citep{sun2018beyond, park2020relation}, semantic parsing \citep{meng2020parsing, kalayeh2018human}, pose estimation \citep{suh2018part, su2017pose}, region of interest \citep{he2019part} and key points \citep{wang2020high, khorramshahi2019dual} to further mine fine-grained information are typical ideas in Re-ID. At the CNN backbone level, some people try to directly improve the convolutional layer and residual block \citep{zhou2019omni}, and a large number of studies introduce attention modules in CNNs to capture the relationship between different convolutional channels, feature maps, and local regions \citep{guo2019two, ye2021deep, li2018harmonious, zhang2020relation, chen2019mixed}. On the other hand, for video sequence input with temporal information, the major limitation of CNNs is that it can only process spatial dimension information. Some video Re-ID works introduce RNN or LSTM for sequence modeling \citep{mclaughlin2016recurrent, yan2016person, liu2017video}. 

Open-world technologies usually target more complex and difficult scenarios, including cross-modal Re-ID, unsupervised learning, domain generalization, \etc. (1) \textit{Cross-modal Re-ID.} In recent years, cross-modal Re-ID of visible-infrared \citep{ye2021channel, li2020multi, yang2023towards, cheng2023unsupervised} and text-image \citep{niu2020improving, wu2021lapscore, ding2021semantically} has received more and more interests. For visible-infrared Re-ID, researchers design single-stream \citep{ye2020visible}, dual-stream \citep{ye2019bi, ye2018hierarchical, zhang2021attend} and other different structures \citep{wu2017rgb} based on CNN to learn modality-sharing and modality-specific feature representations. To reduce the difference between modalities, a class of methods implements modal conversion or style transformation through GAN \citep{wang2019rgb, wang2019learning, wang2020cross} or special augmentation strategies \citep{ye2021channel} and then performs subsequent CNN-based feature representation learning. Text-to-image Re-ID mainly focuses on the cross-modal alignment module design based on the visual and text features extracted from each modality backbone \citep{zhang2018deep, sarafianos2019adversarial}. In addition, many works also introduce attention mechanisms to enhance local information matching \citep{shao2022learning, farooq2022axm}. (2) \textit{Unsupervised learning.}  This approach alleviates the label insufficiency issue \citep{ge2020self}, which now has been a trending topic due to its benefits in large-scale applications. It mainly includes two categories: unsupervised domain adaptation \citep{dai2021idm, bai2021unsupervised, zheng2021group} and pure unsupervised learning \citep{lin2020unsupervised, wang2020unsupervised}. 
In addition to transferring knowledge from labeled source datasets to unlabeled target datasets \citep{wei2018person, deng2018image}, most of the existing methods learn feature representations purely from unlabeled images \citep{zhang2022implicit}. The core idea is to use the features extracted by CNNs to perform clustering to obtain pseudo-labels as label supervision, some of which focus on generating high-quality pseudo-labels \citep{cho2022part, zhang2021refining, wu2022pseudo}, and others improve clustering algorithms and training strategies~\citep{lin2019bottom, ge2020self, dai2022cluster}. (3) \textit{Other open scenes.} 
In recent years, an increasing number of research efforts have shifted towards open scenarios, such as cloth-changing Re-ID and domain-generalizable Re-ID. To facilitate the research, many new cloth-changing datasets \citep{qian2020long, yang2019person} have been introduced. The key to addressing the cloth-changing problem lies in learning clothing-agnostic features, and straightforward approaches involve augmenting the data by introducing a variety of clothing types \citep{jia2022complementary, xu2021adversarial}. Many works try to utilize auxiliary information, such as human parsing \citep{liu2023dual,guo2023semantic}, gait \citep{jin2022cloth}, shape \citep{hong2021fine} to guide the CNN model to focus on identity-related features. Domain generalization is also highly aligned with practical application requirements. Some research endeavors focus on creating large-scale and diverse synthetic data \citep{li2021weperson}, while others seek to enhance the generalization capabilities of CNN models through meta-learning \citep{choi2021meta, ni2022meta} or disentanglement techniques \citep{jin2020style}.

\subsection{Understanding and Analysis of Transformer}
\label{sec:transformer_analysis}
The introduction of Vision Transformer opens novel directions for Re-ID studies, especially in challenging scenarios. In this subsection, we first give the basic concept of the Transformer (\S \ref{sec:transformer_background}). In order to demonstrate the superiority of the Transformer, we provide a comprehensive comparison between the Transformer and CNN and analyze it in terms of network architecture, modeling capabilities, scalability, flexibility, and special properties (\S \ref{sec:transformer_superiority}).

\subsubsection{Transformer Concepts}
\label{sec:transformer_background}
\textbf{Original Transformer.} The original Transformer \citep{vaswani2017attention} was proposed in the field of natural language processing (NLP), which is the first sequence transduction model based on the attention mechanism. It completely abandons the dominant sequence transduction models based on complex recurrent and convolutional neural networks and achieves new state-of-the-art levels in multiple NLP tasks. The transformer is essentially an encoder-decoder structure, in which both the encoder and decoder are composed of multiple stacked transformer layers \citep{vaswani2017attention, han2022survey, xu2023multimodal}. Each transformer layer consists of two sub-layers: a multi-head self-attention mechanism and a position-wise fully connected feed-forward network. Self-attention plays a crucial role in Transformer, which enables each element to learn to gather from other tokens in the sequence. Multi-head self-attention can create multiple attention matrices in a layer, and with multi-head, the self-attention layer will create multiple outputs to guarantee diverse capability. The two sub-layers perform residual connection \citep{he2016deep} for stability, followed by layer normalization. The transformer accepts tokenized sequences as input. To make efficient use of sequence order, an optional positional encoding (relative or absolute) needs to be added. Transformers are used in machine translation tasks, where the encoder extracts features from input with positional encodings, and the decoder uses these features to produce output. Since Transformer was proposed, it has gradually become mainstream and most of the subsequent NLP research has been reprocessed on its basis \citep{devlin2018bert}.


\textbf{Vision Transformer.}
The emergence of Vision Transformer (ViT) \citep{dosovitskiyimage} is a significant breakthrough in the field of computer vision \citep{cheng2023hybrid}. Different from the previous work that embeds the attention module in the CNN, it applies the pure Transformer to the image patches with a simple idea and has achieved remarkable results. Specifically, given an image $\mathbf{x} \in \mathbb{R}^{H \times W \times C}$, $H \times W$ and $C$ represent the image resolution and the number of channels respectively. In order to adapt to the tokenized sequence input of the Transformer, ViT designs a patch embedding operation that divides an image into $N$ patches, where the size of each patch is $P \times P$. These patches are projected into the $D$-dimensional space after linear transformation as the input of ViT, which is a sequence composed of N $D$-dimensional vectors, denoted as $\mathbf{x} \in \mathbb{R}^ {N \times D}$. A special learnable embedding called class token is set for classification, which is directly concatenated to the patch embedding. Following a similar line of the original Transformer, the position embedding is also added to each patch embedding to preserve the spatial position information of the image which is represented as $E_{pos} \in \mathbb{R}^{(N+1) \times D}$. ViT adopts the same structure as the encoder of the original Transformer as a feature extractor. Following the ViT paradigm, a series of subsequent ViT variants are proposed for various vision tasks, leading to significant advancements \citep{han2022survey}.

\subsubsection{Superiority of Vision Transformer}
\label{sec:transformer_superiority}
We provide a detailed analysis of the strengths of Transformer from the vision perspective to facilitate the subsequent elaboration of its robust performance in addressing complex and dynamic Re-ID scenarios.

\textbf{Powerful Modeling Capabilities.}
Different from the standard CNNs, which are limited to the local receptive field, it is extremely difficult to establish long-distance relationships at the early stage. As a result, the performance of CNNs is limited for challenging scenarios. In contrast, the powerful modeling ability of Transformer is reflected in the local-global duality \citep{walmer2023teaching}. Specifically, the modeling of images mainly involves pixel level and object level in vision tasks with images or videos. The attention mechanism of the Transformer is flexibly designed to process information from any image region and can model any relationships between pixels-pixels, objects-pixels and objects-objects. Moreover, CNN can only build hierarchical representations from local to global, whereas Transformer has the flexibility to integrate global information at any stage \citep{naseer2021intriguing, liu2021swin}. For Re-ID, both global and local modeling are essential for learning discriminative features to distinguish high-similar inter-class objects.

\textbf{Diverse Unsupervised Learning Paradigms.}
Due to the expensive and time-consuming nature of acquiring large amounts of high-quality annotated data, unsupervised learning can develop more generalized feature representations without relying on annotations. The great success of Transformers in NLP has largely benefited from self-supervised learning, which provides a solid foundation for the self-supervised research in Vision Transformer. 
Unsupervised learning in the field of computer vision has primarily been centered around contrastive learning, and the introduction of Transformer has made it feasible to incorporate mainstream generative learning approaches from NLP, such as masked autoencoders \citep{he2022masked}. Besides, discriminative self-supervised methods reveal some new characteristics in Transformer models, such as clear object boundaries \citep{caron2021emerging}. In general, unsupervised learning with its cost-effectiveness and generalization capabilities, emerges as a future trend, and transformers hold a unique advantage within this trend.

\textbf{Multi-modal Uniformity and Versatility.}
In practical applications, single-modal data may lack information or be ambiguous. Multi-modal learning allows models to leverage diverse types of data, enabling them to capture rich and comprehensive information for a better understanding of complex real-world scenarios. Compared to CNN, which is primarily designed for processing image modalities, the Transformer exhibits significant versatility across multiple modalities. Multi-modal information can be transformed into a token sequence or latent space features within the same semantic space and input into Transformer for encoding. Transformer can be considered as a fully connected graph, where each token embedding is represented as a node in the graph and the relationships between these embeddings can be described by edges. This property enables Transformer to function within a modality-agnostic pipeline that is compatible with various modalities \citep{xu2023multimodal}. 
Especially in the combination of vision and language, Transformer promotes many new ideas for solving cross-modal Re-ID challenges.

\textbf{High Scalability and Generalization.}
With the continuous increase in data, the future demand for highly scalable models becomes increasingly urgent to adapt effectively to the growing scale of data. Moreover, the generalization capability is crucial for stable performance in dynamic and unknown environments. Numerous recent studies have shown that the powerful scalability of Transformer in terms of large models and big data has achieved incredible results \citep{brown2020language, dehghani2023scaling}. Zhai \textit{et al.} \citep{zhai2022scaling} successfully trained a Vision Transformer model with 2 billion parameters, achieving a new record of 90.45\% top-1 accuracy on ImageNet. Transformers hold immense potential for larger and more versatile models. The encoder-decoder structure of the Transformer, coupled with the joint learning of decoder embeddings and positional encoding, can seamlessly unify various tasks. Additionally, its powerful cross-modal learning capabilities offer further possibilities for expanding Re-ID applications.

\section{Transformer in Object Re-ID}
\label{sec:Transformer Re-ID}
In this section, we comprehensively review the latest research on Transformer-based Re-ID. Considering the different types of challenges and diverse applications of Re-ID tasks, we divided the existing research into four scenarios: regular images or videos with annotations (\S \ref{sec:image/video_Re-ID}), limited data or limited annotations (\S \ref{sec:unsupervised_Re-ID}), multimodal data (\S \ref{sec:mutimodal_Re-ID}), and special settings (\S \ref{sec:special_Re-ID}) to demonstrate the advantages of Transformer respectively.

\subsection{Transformer in Image/Video Based Re-ID}
\label{sec:image/video_Re-ID}
In this subsection, we summarize the progress of transformers under the general supervised setting of image-based (\S \ref{sec:image_Re-ID}) and video-based (\S \ref{sec:video_Re-ID})  Re-ID. For image-based Transformer Re-ID methods, we first review different structural designs at the backbone level for discriminative Re-ID feature extraction. In addition, the tokenized embeddings and attention mechanism in Transformer provide strong flexibility for representation learning. We comprehensively summarize the methods of exploiting Transformer properties for Re-ID-specific design. In video Re-ID, Transformer-based methods have shown great superiority over CNNs on modeling the spatio-temporal cues.

\begin{figure}
    \centering
    \includegraphics[width=\linewidth]{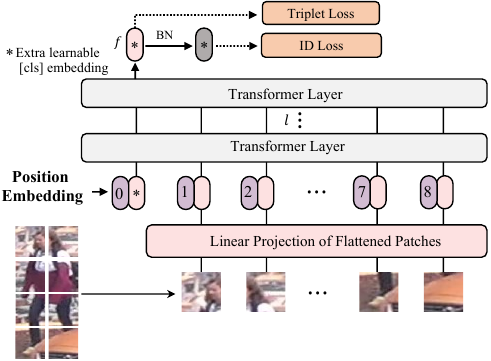}
    \caption{The first pure transformer baseline for object Re-ID \citep{he2021transreid}. The Vision Transformer backbone \citep{dosovitskiyimage} is adopted as a feature extractor, optimized with ID loss and triplet loss \citep{arxiv17triplet} widely used in Re-ID.}
    \label{fig:transreid}
    \vspace{-3mm}
\end{figure}

\subsubsection{Transformer in Image-based Re-ID}
\label{sec:image_Re-ID}

\begin{figure*}[t]
    \centering
    \includegraphics[width=\linewidth]{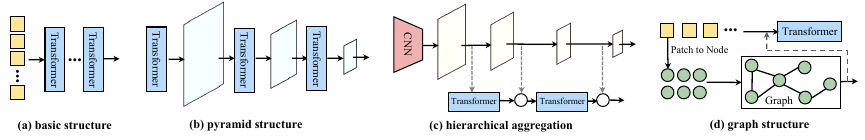}
     \vspace{-3mm}
    \caption{\textbf{Different Transformer architectures designed for image-based Re-ID.} (a) The basic Re-ID baseline based on Vision Transformer \citep{he2021transreid}. (b) Pyramid Transformer for learning multi-scale features \citep{li2022pyramidal}. (c) Transformer and CNN hybrid architecture for aggregating hierarchical features \citep{zhang2021hat}. (d) Combination of graph structure and Transformer \citep{shen2023git}. }
     \vspace{-2mm}
    \label{fig:sturctrue}
\end{figure*}

\textbf{Architecture Improvements.} A number of recent studies have shown that applying Vision Transformer as a feature extractor in Re-ID can achieve high accuracy \citep{he2021transreid, cao2022pstr, zhang2022interlaced, li2023dc}. Many recent Re-ID methods are dedicated to designing special Transformer architectures to build stronger backbones \citep{li2022pyramidal, shen2023git, zhang2021hat}. The first to introduce Vision Transformer in the Re-ID field is TransReID \citep{he2021transreid}, which preserves two advantages of Transformer at the architectural level. Compared with CNNs, the multi-head self-attention scheme in Transformer captures long-distance dependencies so that different object/body parts can be better focused. In addition, Transformer retains more detailed information without down-sampling operators. Therefore, it builds a pure Transformer baseline for supervised image-based single modality object Re-ID, following in a similar way to ViT \citep{dosovitskiyimage}, as shown in Fig. \ref{fig:transreid}. Even simply replacing the feature extraction network in basic Re-ID methods with vision transformers, the performance on multiple vehicle and person Re-ID datasets is comparable to state-of-the-art methods, reflecting the strong potential of Transformers for Re-ID tasks. Inspired by this, some subsequent methods design special transformer architectures such as pyramid structure \citep{li2022pyramidal}, hierarchical aggregation \citep{zhang2021hat, tan2023mfat}, graph structure \citep{shen2023git}, \etc, while some other methods intent to improve the attention mechanisms \citep{chen2021oh, zhu2022dual, tian2022hierarchical, shen2023x}. 

Considering the mutual cooperation of CNN and Transformer, Li \textit{et al.} \citep{li2022pyramidal} develop a pyramidal transformer structure like CNN to learn multi-scale features and improve the patch embedding process, utilizing convolution with the anti-aliasing block to capture translation-invariant information. Similarly, based on hierarchical features extracted by CNN, HAT \citep{zhang2021hat} is proposed to aggregate features of different scales in a global view with the help of Transformer. GiT \citep{shen2023git} introduces graphs in transformers to mine relationships of nodes within the patch. 

For the improvement of attention schemes in Transformer, Zhu \textit{et al.} \citep{zhu2022dual} present a dual cross-attention learning strategy by emphasizing the interaction between the global image and local high-response regions and the interaction between image pairs. Considering the impact of variable appearances of the same identity, Shen \textit{et al.} \citep{shen2023x} introduce cross-attention in the Transformer encoder to merge information from different instances. From the perspective of improving Transformer efficiency in Re-ID, Tian \textit{et al.} \citep{tian2022hierarchical} present a hierarchical walking attention, by introducing a prior as an indicator to decide whether to skip or calculate a region's attention matrix in the image patch. For lightweight Re-ID models, Mao \textit{et al.} \citep{mao2023attention} design an attention map-guided Transformer pruning method, so that the models can be deployed on edge devices with limited resources. It removes redundant tokens and heads in a hardware-friendly manner, achieving the goal of reducing the inference complexity and model size without sacrificing the accuracy of Re-ID.


\textbf{Re-ID-specific Design.}
For different objects and Re-ID-specific challenges, many works explore the application of Transformer to make Re-ID-specific adaptations \citep{arXiv2021AAformer, lai2021transformer, li2021diverse}. For the most crucial local discriminative information mining for Re-ID, Vision Transformer naturally has attention and patch embeddings, which can be easily used to capture local discriminative information to enhance the representation \citep{he2021transreid, li2021diverse, lai2021transformer, arXiv2021AAformer}. Furthermore, the disentanglement of some key information can be modeled by the encoder-decoder structure in Transformer \citep{li2021diverse, wang2022pose, zhou2022motion}. Besides, the structure prior \citep{chen2022sketch} or task specialties \citep{chen2022rotation} of different objects are also important for Transformer design. 

The effectiveness of learning local feature representation has been proven by extensive Re-ID research, and adding an attention mechanism to the CNN to focus on more discriminative information is a mainstream practice in previous studies (\S \ref{sec:Re-ID_methods}). However, many new ideas that use the special properties of Transformers to learn local features are also emerging and growing rapidly \citep{qian2022unstructured}. TransReID \citep{he2021transreid} designs shift and patch shuffling operations on the transformer baseline to learn local features, which is conducive to enhancing disturbance invariance and robustness. Zhu \textit{et al.} developed an Auto-Aligned Transformer (AAformer) \citep{arXiv2021AAformer} to adaptively locate the human parts. It learns local representations by introducing learnable part tokens to the transformer and integrates part alignment into self-attention. Zhang \textit{et al.} revealed that self-attention leads to the inevitable dilution of high-frequency components of images. To enhance the feature representation of high-frequency components that are important to Re-ID, they proposed to use Discrete Haar Wavelet Transform (DHWT) \citep{mallat1989theory} to split the patches of high-frequency components as the auxiliary information \citep{zhang2023pha}. \blue{ For vehicle Re-ID, feature misalignment caused by pose and viewpoint variations is a key challenge. Previous methods addressed this by aligning features based on additional annotations of vehicle parts. To make feature decomposition more flexible and unstructured, \citep{qian2022unstructured} leverages Transformers to decouple vehicle features across the spatial dimension, enabling fine-grained feature learning on a global scale. Moreover, Transformers are highly effective in establishing interactions between semantic knowledge related to Re-ID and visual features. MsKAT \citep{li2022mskat} introduces a state elimination Transformer to remove interference from cameras and viewpoints, as well as an attribute aggregation Transformer to gather information on vehicle attributes such as color and type. }

\begin{table*}[t]
\renewcommand{\arraystretch}{1.5}
    \centering
    \caption{\label{tab:image/vidoe} Representative Transformer methods based on image/video Re-ID. }
    \resizebox{\textwidth}{!}{
    \begin{tabular}{c|ccccc}
    \hline
        \textbf{Category} & \textbf{Focal Point} & \textbf{Object} & \textbf{Transformer Type} & \textbf{Method} &\textbf{Publication} \\ \hline
        \multicolumn{6}{c}{\textbf{Image-based Re-ID}}\\ \hline
        \multirow{5}{*}{\shortstack{Architecture \\ Design}} & Pure Transformer Re-ID baseline & Vehicle\&Person & ViT & TransReID \citep{he2021transreid} & ICCV \\ 
         & Hierarchical feature aggregation & Person & Hybrid & HAT \citep{zhang2021hat} & ACM MM  \\ 
         & Introducing the benefits of CNN & Person & PVT & PTCR \citep{li2022pyramidal} & ACM MM  \\ 
         & Improvements to attention  & Vehicle\&Person & ViT,DeiT & DCAL \citep{zhu2022dual} & CVPR  \\ 
         & Integrating graph structure & Vehicle & ViT & GiT \citep{shen2023git} & TIP  \\ 
        \hline
        \multirow{6}{*}{\shortstack{Re-ID-specific \\ Design}} & Partial representation learning & Person & Decoder & PAT \citep{li2021diverse} & CVPR \\ 
         & Introducing auxiliary information & Person & Encoder-Decoder & PFD \citep{wang2022pose} & AAAI \\
         & Modeling relationships between individuals & Person & Transformer & NFormer \citep{wang2022nformer} & CVPR  \\
         & Learning rotation-invariant features  & Vehicle\&Person & ViT & RotTrans \citep{chen2022rotation} & ACM MM \\
         & High-frequency augmentation & Person & ViT & PHA \citep{zhang2023pha} & CVPR \\
         \hline 
        \multicolumn{6}{c}{\textbf{Video-based Re-ID}} \\ \hline
        \multirow{2}{*}{\shortstack{Combination \\ of CNN \& \\Transformer}}&Transformer for post-processing & Person & Decoder & DenseIL\citep{he2021dense} & ICCV \\ 
        & Coupled CNN-Transformer & Person & ViT/Swin/DeiT & DCCT\citep{liu2023deeply} & TNNLS \\ \hline
        \multirow{2}{*}{\shortstack{Pure \\Transformer}} & Spatial-temporal aggregation & Person & ViT & MSTAT \citep{tang2022multi} & TMM  \\ 
        & Spatial-temporal joint modeling & Person & ViT & CAViT \citep{wu2022cavit} & ECCV \\ \hline
    \end{tabular}}
\end{table*}

\subsubsection{Transformer in Video-based Re-ID}
\label{sec:video_Re-ID}
\textbf{Transformer for Post-processing.} Video Re-ID aims to fully exploit the temporal and spatial interactions of frame sequences to extract more discriminative representations \citep{wu2022cavit}. Compared with CNN-based methods that require additional models to encode time information, Transformer is proposed as a powerful architecture for processing sequence data, which has inherent advantages. The global attention mechanism in Transformer can be easily adapted to video data to capture spatio-temporal dependencies \citep{tang2022multi}. A group of Transformer-based video Re-ID methods are hybrid architectures \citep{liu2021video, zhang2021spatiotemporal, he2021dense}. 
They typically refer to processing the extracted features from other models (such as convolutional neural networks) before further processing them using a Transformer model. The primary use of Transformer lies in its self-attention mechanism, which captures long-term dependencies and contextual information within the sequence.
Zhang \textit{et al.} \citep{zhang2021spatiotemporal} designed a two-stage spatio-temporal transformer module for patch tokens converted from CNN feature maps, where the spatial transformer focuses on object regions with different backgrounds, while the subsequent temporal transformer focuses on video sequences to exclude noisy frames. Also based on the features extracted by CNN, Liu \textit{et al.} \citep{liu2021video} present a multi-stream Transformer architecture that emphasizes three perspectives of video features via spatial Transformer, temporal Transformer, and spatio-temporal Transformer. A cross-attention based strategy is designed to fuse multi-view cues to obtain enhanced features. Additionally, some studies consider the complementary learning of CNN and Transformer in space and time. Specifically, DCCT \citep{liu2023deeply} introduces self-attention and cross-attention to the features extracted by two separate networks to establish a spatial complementary relationship, and designs hierarchical aggregation based on a temporal Transformer to integrate two temporal features. DenseIL \citep{he2021dense} is a hybrid architecture consisting of a CNN encoder and a Transformer decoder with dense interaction, where the CNN encoder extracts discriminative spatial features while the decoder aims to densely model spatio-temporal interactions across frames.

\textbf{Pure Transformer.} The hybrid architecture makes it difficult to overcome the intrinsic limitations of CNN for perceiving long-distance information. Some recent work attempts to explore the application of pure Transformer architecture to video Re-ID \citep{tang2022multi, wu2022cavit}. Tang \textit{et al.} \citep{tang2022multi} designed a multi-stage Transformer framework by taking advantage of Vision Transformer's class token to facilitate the aggregation of various information. At different stages, the learning of attribute-associated information, identity-associated information and attribute-identity-associated information is guided respectively. Besides, different from the mainstream divide-and-conquer strategy that tackles feature representation and feature aggregation separately that fail to simultaneously solve temporal dependence, attention and spatial misalignment, a contextual alignment Vision Transformer (CAViT) \citep{wu2022cavit} is proposed for spatial-temporal joint modeling. To jointly model spatio-temporal cues, it replaces self-attention with temporal-shift attention based on a pure Transformer architecture to align objects in adjacent frames. \blue{
In video person Re-ID, occlusion remains a major challenge, as traditional convolution-based methods often struggle to effectively handle occlusion and the misalignment of adjacent frames, leading to a drop in recognition performance. To address this issue, TCViT \citep{wu2024temporal} leverages Transformers, utilizing their attention mechanisms to focus on the relative motion and completeness of frame-level features, aligning the frames and improving the visibility of the target person. This approach significantly enhances the model's ability to handle occlusion.}

\subsection{Transformer in ReID with Limited Data/Annotations}
\label{sec:unsupervised_Re-ID}
In this survey, limited annotation usually corresponds to unsupervised learning Re-ID technology (\S \ref{sec:usl_Re-ID}), while limited data mainly focuses on domain generalization in Re-ID (\S \ref{sec:dg_Re-ID}). In fact, Transformer is still in the preliminary exploration of such Re-ID scenarios, with a small number of works but showing great potential.

\begin{table*}[t]
\renewcommand{\arraystretch}{1.5}
    \centering
    \caption{\label{tab:usl-performance} Comparison of state-of-the-art supervised and unsupervised methods based on CNN and Transformer on two widely used datasets Market-1501 \citep{zheng2015scalable} and MSMT17 \citep{wei2018person} in Person Re-ID. The performance of different pre-training conditions is reported. The supervised TransReID-SSL results are obtained by basic Transformer baseline \citep{he2021transreid} fine-tuning and the unsupervised TransReID-SSL results are obtained by Cluster-Contrast \citep{dai2022cluster} fine-tuning. TransReID-SSL* refers to the results reproduced as a baseline in our experiments.}
     \vspace{-3mm}
    \resizebox{\textwidth}{!}{
    \begin{tabular}{c|cccccccc}
    \hline
       ~ & ~ &~ &\multicolumn{2}{c}{ Pre-training Conditions} & \multicolumn{2}{c}{Market-1501} & \multicolumn{2}{c}{ MSMT17} \\ \hline
        Method & Venue & Backbone & Data & Supervision & mAP & Rank-1 & mAP & Rank-1 \\ \hline
        \multicolumn{9}{c}{\textbf{State-of-the-art methods for supervised Re-ID}}\\ \hline
        AGW \citep{ye2021deep} & TPAMI  & CNN & ImageNet & Supervised & 87.8 & 95.1 & 49.3 & 68.3 \\ 
        CDNet \citep{li2021combined} & CVPR  & CNN & ImageNet & Supervised & 86.0 & 95.1 & 54.7 & 78.9 \\ \hline
        TransReID \citep{he2021transreid} & ICCV  & Transformer & ImageNet & Supervised & 89.5 & 95.2 & 69.4 & 86.2 \\ 
        PHA \citep{zhang2023pha} & CVPR & Transformer & ImageNet & Supervised & 90.2 & 96.1 & 68.9 & 86.1 \\ 
        TransReID-SSL \citep{luo2021self}& Arxiv  & Transformer & LUPerson & SSL & \textbf{93.2} & \textbf{96.7} & \textbf{75.0} & \textbf{89.5} \\ \hline
        \multicolumn{9}{c}{\textbf{State-of-the-art methods for unsupervised Re-ID}}\\ \hline
        ICE \citep{chen2021ice} & ICCV  & CNN & ImageNet & Supervised & 82.3 & 93.8 & 38.9 & 70.2 \\ 
        ISE \citep{zhang2022implicit} & CVPR  & CNN & ImageNet & Supervised & 85.3 & 94.3 & 37.0 & 67.6 \\ 
        Cluster-Contrast \citep{dai2022cluster} & ACCV  & CNN & ImageNet & Supervised & 83.0 & 92.9 & 31.2 & 61.5 \\ 
        Cluster-Contrast \citep{dai2022cluster} & ACCV  & CNN & LUPerson & SSL & 84.0 & 94.3 & 31.4 & 58.8 \\ \hline
        PASS \citep{zhu2022pass} & ECCV  & Transformer & LUPerson & SSL & 88.5 & 94.9 & 41.0 & 67.0 \\ 
        TransReID-SSL \citep{luo2021self} & Arxiv  & Transformer & LUPerson & SSL & 89.6 & 95.3 & 50.6 & 75.0 \\ \hline
        TransReID-SSL* \citep{luo2021self} & Arxiv & Transformer & LUPerson & SSL & 89.9 & 95.2 & 48.2 & 72.8 \\
       UntransReID (Ours)  & - & Transformer & LUPerson & SSL & \textbf{90.7} & \textbf{95.7} & \textbf{51.1} & \textbf{75.7} \\ \hline
    \end{tabular}
    }
\end{table*}

\subsubsection{Transformer in Unsupervised Re-ID}
\label{sec:usl_Re-ID}

\textbf{Self-supervised Pre-training.} Generally, the existing unsupervised Re-ID methods mainly rely on the features extracted by CNN to cluster and generate pseudo-labels as label supervision, in which CNN is supervisedly pre-trained on ImageNet \citep{zhang2022implicit, dai2022cluster}. However, supervised pre-training focuses on coarse category-level distinction, which reduces the rich visual fine-grained information in images. These fine-grained cues are crucial for Re-ID tasks with a large intra-class variation. A class of studies of Transformer in unsupervised Re-ID emphasizes 
self-supervised pre-training to obtain a better initialization model and reduce the domain difference between ImageNet data and Re-ID data \citep{zhu2022pass, luo2021self}. The success of self-supervised Transformers in vision tasks provides a lot of modeling and training experience for unsupervised Re-ID research \citep{han2022survey}. The advantages of Vision Transformer in unsupervised learning are reflected in several aspects: 1) The strong scalability of the Transformer model for large-scale unlabeled data. Self-supervised learning  can fully use the representation ability of the models with Transformer architecture \citep{caron2021emerging}. 2) The flexibility of the Transformer structure provides more diverse self-supervised paradigms, which are extremely challenging for CNNs to complete \citep{he2022masked}. 

With the emergence of LUPerson \citep{fu2021unsupervised}, a large-scale unlabeled dataset specifically for person Re-ID, Luo \textit{et al.} \citep{luo2021self} began to initially explore effective Transformer self-supervised pre-training paradigms for Re-ID, achieving significant results. They first conduct extensive experiments to investigate the performance of CNNs and Transformers using different Self-Supervised Learning (SSL) methods on ImageNet and LUPerson pre-training datasets. In addition, they promote the stability and domain invariance of Transformer by designing the IBN-based convolution stem to replace the standard patchify stem in ViT to enhance the local feature learning. The conclusion is that Transformer is ahead of CNN in terms of pre-training. Notably, under the fully unsupervised condition of using DINO \citep{caron2021emerging} to pre-train the Transformer on LUPerson and fine-tuning with a common unsupervised Re-ID method \citep{dai2022cluster}, the performance of Re-ID is even competitive with the state-of-the-art supervised Re-ID method. It can be regarded as a major breakthrough in the field of unsupervised Re-ID. On this basis, the later PASS \citep{zhu2022pass} further integrated Re-ID-specific part-aware properties in the self-supervised Transformer pre-training. Inspired by DINO \citep{caron2021emerging}, which develops a simple strategy for label-free self-distillation, PASS introduces several learnable tokens to extract part-level features, further reinforcing fine-grained learning for Re-ID. Specifically, it divides the image into several fixed overlapping local regions and randomly crops local views from them, while the global view is randomly cropped from the whole image. In knowledge distillation, all views are passed through the student and only the global view is passed through the teacher.

The pre-trained Transformer serves as a powerful initialization model compared with previously widely-used ImageNet pretraining. It can be fine-tuned with different Re-ID supervised learning or unsupervised learning methods in downstream tasks. As shown in Table \ref{tab:usl-performance}, the performances of corresponding methods have been greatly improved. We believe these research efforts will be an advancement for the Re-ID community, allowing future work to be performed with better pre-trained models.

\textbf{Unsupervised Domain Adaptation.} Transformer has received limited attention for another widely studied unsupervised domain adaptation (UDA) problem in unsupervised Re-ID, with a small amount of work on vehicles and persons respectively \citep{wang2022body, wei2022transformer}. Different from the previous Re-ID method based on domain alignment to guide feature learning to achieve distribution consistency at the domain level or identity level, Wang \textit{et al.} \citep{wang2022body} oriented person Re-ID to achieve fine-grained domain alignment between different body parts with the help of Transformer. They embed the transformer layer into the feature extraction backbone and discriminators respectively, where the backbone obtains the class tokens representing each body part and the discriminators extract the domain information contained in each body part. Dual adversarial learning is introduced in the backbone and discriminator to align each class token of a target domain sample with the corresponding class token in the source domain. Another vehicle-oriented Transformer work belongs to the clustering-based UDA solution \citep{wei2022transformer}. The core idea is to make the Transformer adaptively focus on the discriminative part of the vehicle in each domain through a joint training strategy. To achieve dynamic knowledge transfer, both source and target images are simultaneously fed into a shared CNN to obtain feature maps, and the Transformer encoder-decoder architecture is subsequently introduced to generate a global feature representation integrating contextual information from the feature maps. Based on Transformer, a learnable domain encoding module similar to positional encoding is added to better utilize the specific characteristics of each domain.

\subsubsection{Transformer in Generalized Re-ID}
\label{sec:dg_Re-ID}

The application of the Transformer promotes new ideas of Re-ID in the challenging problem of domain generalization (DG) \citep{liao2021transmatcher}. Completely different from mainstream research that uses Transformer for feature representation learning, TransMatcher \citep{liao2021transmatcher} studies Transformer for image matching and metric learning for a given image pair from the perspective of generalizability. For Re-ID, a typical image matching and metric learning problem, the Transformer encoder can only facilitate the feature interaction between different positions within an image but fails to realize the interaction between different images. Liao \etal \citep{liao2021transmatcher} also demonstrate that directly applying vanilla vision Transformer ViT through  a classification training pipeline will result in poor generalization to different datasets. Inspired by the cross-attention module in the Transformer decoder that enables cross-interaction between query and encoded memory, they attempt to use actual image queries instead of learnable query embeddings as the input to the decoder to gather information across query-key pairs, effectively boosting performance. Further, TransMatcher is designed as a simplified decoder more suitable for image matching, which discards all attention implementations with softmax weighting and only keeps query-key similarity computation. This study demonstrates that Transformer can be effectively adapted to image matching and metric learning tasks with strong potential, and now it has been widely used in later research \citep{ni2023part,wang2023few} to improve the generalizability.

In addition, researchers try to investigate the generalization ability of Transformer in Re-ID. Ni \textit{et al.} \citep{ni2023part} employed different Transformers and CNNs as the backbone to assess the cross-domain performance from Market \citep{zheng2015scalable} to MSMT \citep{wei2018person}. The results indicate that Vision Transformers outperform CNNs significantly. 
On this basis, a proxy task is introduced which mines local similarities shared by different IDs based on part aware attention, to promote the Transformer to learn generalized features without using ID annotations.

\blue{
In terms of generalization, Transformers have shown great promise by effectively focusing on the object of interest and learning domain-invariant features that can transfer well across different environments. This ability to capture robust, generalized features is one of the key strengths of Transformers, making them suitable for complex Re-ID tasks across various settings. However, when it comes to generalizing across different types of objects, the limitations of Transformers become more apparent. Their attention mechanism may struggle to adapt to significant variations between object categories, particularly when the intra-class variations are subtle, but the inter-class differences are substantial.}

\subsection{Transformer in Cross-modal Re-ID}
\label{sec:mutimodal_Re-ID}

In this subsection, we summarize the Transformer progress of three types of cross-modal problems that have received more attention in Re-ID: visible-image (\S \ref{sec:visible-infrared}), text-image (\S \ref{sec:text-image}) and sketch-image (\S \ref{sec:sketch-image}). Recently, Transformer has made a lot of novel works and influential breakthroughs in multi-modal learning in the field of vision. The primary advantage of the Transformer is that its input can include multiple sequences of tokens, each with distinct attributes, facilitating the association of different modalities via the attention mechanism without necessitating any changes to the architecture \citep{xu2023multimodal}.

\subsubsection{Visible-Infrared Re-ID}
\label{sec:visible-infrared}
Visible-infrared Re-ID is a cross-modal retrieval task that aims at matching the daytime visible and nighttime infrared images \citep{wu2017rgb,ye2019improving}. 
The major challenge of visible-infrared Re-ID is the modality gap between two types of images.
The application of the Transformer provides many benefits to the visible-infrared Re-ID problem. For example, Transformers tend to learn shape and structure information, while CNNs rely on local texture information \citep{naseer2021intriguing}. Due to the lack of colors and lighting conditions in infrared images, Vision Transformer can better capture modality-invariant information and has stronger robustness. On the other hand, the vision transformer structure and attention enable local cross-modal associations to be easily established at the patch token level, which is essential for fine-grained properties of Re-ID, especially under a large modality gap.

The mainstream approach in existing visible-infrared Re-ID is to learn modality-shared features, which decouple features into modality-specific and shared-modal features, and then focus on modality alignment at the feature level.
Jiang \textit{et al.} \citep{jiang2022cross} started trying to adopt Transformer's encoder-decoder architecture for modal feature enhancement and compensation to promote better alignment of RGB and IR modalities. They separately construct two sets of learnable prototypes for RGB and IR modalities to represent global modality information. In the Transformer decoder, the IR prototype is regarded as a query for the RGB modality and the part features at the token level of the RGB samples are used as keys and values. Compensated IR modal features are obtained by aggregating partial features through the correspondence of cross-attention between partial features and modal prototypes. In contrast, Liang \textit{et al.} \citep{liang2023cross} introduce learnable embeddings to mine modality-specific features in Transformer in a manner similar to positional encoding, and employ a modality removal process to subtract the learned modality-specific features. 

Considering the specificity of the body part position of the person object, Chen \textit{et al.} \citep{chen2022structure} believe that position interaction can discover the underlying structural relationship between regions and provide more stable invariance for pose changes. A structure-aware position transformer (SPOT) is proposed to extract modality-shared representations. It exploits the attention mechanism to learn structure-related features guided by human key points and adaptively combines partially recognizable cues by modeling context and position relations through a transformer encoder. Additionally, Feng \textit{et al.} \citep{feng2022visible} focus on the interaction of local features across modalities, where they leverage attention to enrich the feature representation of each patch token by interacting with patch tokens from other modalities. Yang \textit{et al.} \citep{yang2023top} argue that each token in the self-attention mechanism in ViT is connected to a class token, where the attention score can be intuitively interpreted as a measure of token importance. To better align features of different modalities, they select top-k important visual patches from each attention head for localizing important image regions. Focusing on the modal invariant information of shallow features such as texture or contour information, Zhao \textit{et al.} \citep{zhao2022spatial} utilize Transformer to encode the spatial information of each convolution stage of CNNs to fuse shallow and deep features to enhance the representation.

\begin{table}[t]
\centering
\caption{\label{tab:clip-reid} \blue{Comparison of the state-of-the-art image-text Re-ID methods based on Transformer.}}
\resizebox{0.5\textwidth}{!}{
\begin{tabular}{c|cc|cc}
\hline
- & \multicolumn{2}{c|}{CUHK-PEDES} & \multicolumn{2}{c}{ICFG-PEDES} \\ \hline
Method & mAP & R1& mAP & R1 \\ \hline
\multicolumn{5}{c}{\textbf{w/o CLIP}} \\ \hline
LGUR \citep{shao2022learning} & - & 65.3 & -& 59.0 \\ 
IVT \citep{shu2022see}  & - & 65.7 & -& 56.0 \\
UniPT \citep{shao2023unified}  & - & 68.5 & -& 60.1 \\\hline
\multicolumn{5}{c}{\textbf{w/ CLIP}} \\ \hline
TP-TPS \citep{wang2023exploiting} & 66.3 & 70.2 & 42.8& 60.6 \\
UNIReID \cite{chen2023towards} & - & 68.7 & -& 61.3 \\
C-Fine\citep{yan2022clip} & - & 69.6 & -& 60.8 \\
IRRA \citep{jiang2023cross}& 66.1 &73.4 &38.1 & 63.5 \\
TBPS-C \citep{cao2024empirical} & 65.4 &73.5 &39.8 & 65.1 \\
MALS \citep{yang2023towards2}  & 66.6 &74.1 &38.9 & 64.4 \\
MLLM\citep{tan2024harnessing} & 69.6 &76.8 &41.5 & 67.0 \\
\hline
\end{tabular}}
\end{table}

\subsubsection{Text-Image Re-ID} \label{sec:text-image}
Text-Image Re-ID refers to a cross-modal retrieval task, which aims at identifying the target object (person or vehicle) from an image gallery based on a given textual query, describing the target appearance \citep{ye2015specific,li2017person}.

\textbf{CLIP in Re-ID.} As a milestone work of Transformer in multimodal applications, the proposal of Contrastive Language-Image Pre-training (CLIP) \citep{radford2021learning} opened up a new era of large-scale pre-training for text-image communication. CLIP uses text information to supervise the self-training of vision tasks, which turns a classification task into an image-text matching task. During the training process, a two-stream network, including an image encoder and a text encoder processes text and image data respectively, and contrastive learning is used to learn the matching relationship between text-image pairs. The pre-trained model directly performs zero-shot image classification without any training data, achieving comparable supervised accuracy. Recently, CILP has become a powerful tool for downstream text-image Re-ID tasks. Some Re-ID works directly introduce CLIP as a pre-training model with good generalization \citep{han2021textbased} or further expand the design of cross-modal association mining \citep{jiang2023cross, yan2022clip, zuo2023plip}, and some works focus more on the utilization of Re-ID related textual information in CLIP to better assist downstream Re-ID tasks \citep{li2023clip, wang2023exploiting}.

Considering the effectiveness of directly fine-tuning CLIP, Yan \textit{et al.} \citep{yan2022clip} explore the transfer of CLIP models to text-image Re-ID. Based on the pre-trained CLIP, they further capture the relationship between image patches and words to build fine-grained cross-modal associations. Inspired by CLIP, Zuo \textit{et al.} \citep{zuo2023plip} propose a language-image pre-training framework PLIP that is more suitable for person objects. To explicitly establish fine-grained cross-modal relations, a large-scale person dataset constructed with stylish generated text descriptions is proposed and three pretext tasks are introduced. The first is semantic-fused image coloring, which recovers the color information of gray-scale person images given a textual description. The second is visual-fused attribute prediction, which predicts masked attribute phrases in text descriptions through paired images. The last is visual-language matching. Instead, with CLIP as the initialization model, IRRA \citep{jiang2023cross} designs a cross-modal implicit relation reasoning module to efficiently construct the relation between visual and textual representations through self-attention and cross-attention mechanisms. This fused representation is used to perform masked language modeling (MLM) task without any additional supervision and inference costs, achieving the purpose of effective inter-modal relation learning. \blue{ He \textit{et al.} \citep{he2023vgsg} developed a CLIP-driven framework focusing on fine-grained cross-modal feature alignment. They proposed a Vision-Guided Semantic Grouping Network, which mitigates the misalignment of fine-grained cross-modal features by semantically grouping textual features and aligning them with visual concepts. Additionally, Wang \textit{et al.} \citep{wang2023exploiting} aim to enhance text-based person search by leveraging the dual generalization capabilities of Vision-Language Pre-training (VLP) models. The paper focuses on fully exploring the potential of textual representations, utilizing pre-trained transferable knowledge in text, and proposes two strategies tailored to descriptive corpora.}
Compared with the previous method which utilizes single-modal pre-trained external knowledge and lacks multi-modal corresponding information, these CLIP-based text-image Re-ID methods have achieved significant performance improvements.

\blue{In addition to text-image Re-ID, some works also leverage CLIP to provide text-based auxiliary information to enhance image-based Re-ID \citep{li2023clip, yang2024pedestrian}.} Compared with the one-hot label of image classification, CLIP-Re-ID \citep{li2023clip} demonstrates that more detailed image text descriptions can help the visual encoder learn better image features, especially for fine-grained tasks such as Re-ID that lack precise descriptions.
Inspired by the learnable prompt used by CoOp \citep{zhou2022learning}, CLIP-Re-ID designs a two-stage training strategy. It combines ID-specific learnable tokens to give ambiguous textual descriptions in the first stage and these tokens together with the text encoder provide constraints for optimizing the image encoder in the second stage. 
\blue{Building on this, Yang \textit{et al.} \citep{yang2024pedestrian} argue that predefined soft prompts may not be sufficient to capture the full visual context and are difficult to generalize to unseen categories. They design an end-to-end PromptSG framework, instead of a two-stage learning process, to leverage CLIP's inherent rich semantics. By utilizing an inversion network to learn representations of specific, more detailed appearance attributes, the framework can create more personalized descriptions for individuals, further enhancing image-based Re-ID.}

\blue{Considering that manually annotating textual descriptions limits the dataset scale, Tan \textit{et al.} \citep{tan2024harnessing} are the first to explore the transferable text-to-image ReID problem. Specifically, they leverage large-scale training data obtained through multimodal large language models (MLLM) to train the model and directly deploy it for evaluation across various datasets. They propose a method to build large-scale datasets with diverse textual descriptions by using MLLM's multi-turn dialogues to generate captions for images based on various templates. To address the issue of MLLM potentially generating incorrect descriptions, they leverage a Transformer-based approach to automatically identify words in the description that do not correspond to the image. This is achieved by analyzing the similarity between the textual content and all patch token embeddings within the image.}


\subsubsection{Sketch/Skeleton Re-ID}
\label{sec:sketch-image}
Sketch-to-photo Re-ID represents a cross-modal matching problem whose query sets are sketch images provided by artists or amateurs \citep{yang2019person}, while the query images in skeleton Re-ID are generated by pose estimation \citep{rao2023transg, rao2024hierarchical}. These two tasks share similar spirits in large information asymmetry.

\textbf{Sketch-image Re-ID.} The significant difference in region-level information between sketches and images is a challenge due to the abstraction and iconography of sketch images. The correlation between object shape and local information plays an important role in sketch-image Re-ID. The advantage of Transformer in learning global-level feature representations shows excellent discriminative ability in sketch photo recognition \citep{chen2022sketch, hong2021fine}. Chen \textit{et al.} \citep{chen2022sketch} experimentally verified that the method using ViT as the backbone has a significant performance improvement over most CNN-based sketch-image Re-ID methods. Therefore, they construct a strong baseline based on a vision transformer for sketch-image Re-ID. In order to narrow the gap between sketches and images, Zhang \textit{et al.} \citep{zhang2022cross} designed a token-level cross-modal exchange strategy in Transformer under the guidance of identity consistency to learn modality-compatible features. Local tokens of different modalities are classified into different groups and assigned specific semantic information to construct a semantically consistent global representation.

In particular, a new, challenging, and modality-agnostic person Re-ID problem has recently been proposed \citep{chen2023towards}. It comprehensively considers descriptive queries such as supplementary text or sketch modalities for general images to achieve multi-modal unified re-identification. Benefiting from CLIP, UNIRe-ID \citep{chen2023towards} developed a simple dual-encoder transformer architecture for multimodality feature learning and designed a task-aware dynamic training strategy that adaptively adjusts the training focus according to the difficulty of the task. This work demonstrates the power of Transformer in multi-modal learning and also opens up the direction for the future promotion of Re-ID.

\textbf{Re-ID with Skeleton Data.}
Person Re-ID via 3D skeletons differs from traditional Re-ID, which relies on visual appearance features such as color, outline, etc., and mainly utilizes the 3D locations of key body joints to model unique body and motion representations. TranSG \citep{rao2023transg} is proposed as a general Transformer paradigm for learning feature representations from skeleton graphs for person Re-ID. Its core idea is to model the 3D skeleton as a graph and use Transformer for full-relational learning of body joint nodes, which simultaneously aggregates key relationship features of body structure and motion into a graph representation. 

\blue{\textbf{Discussion.} For cross-modal Re-ID, Transformer-based methods combined with large language models have made significant progress, particularly in image-text Re-ID, where many breakthroughs have been achieved. However, for other cross-modal Re-ID tasks, such as visible-infrared Re-ID, the challenges remain more pronounced. While Transformers excel at capturing complex relationships between modalities, cross-modal learning in such tasks typically requires extensive amounts of paired data to learn effective feature alignment across domains. Unfortunately, such large-scale datasets are often unavailable or difficult to collect, limiting the scalability and generalization of these methods.}

\subsection{Transformer in Special Re-ID Scenarios}
\label{sec:special_Re-ID}
We investigate that the vision Transformer is also applied to some more open and complex task settings in Re-ID. The more special Re-ID types than the scenarios mentioned above are summarized in this subsection for discussion.

\begin{figure*}[t]
    \centering
    \includegraphics[width=\linewidth]{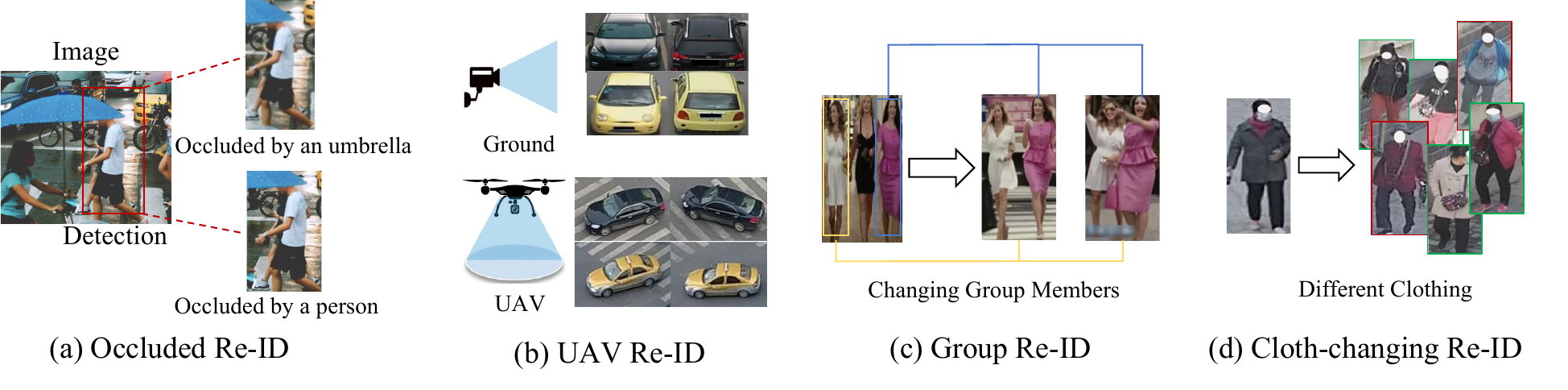}
     \vspace{-3mm}
    \caption{\blue{Description of the characteristics of special Re-ID scenarios.}}
     \vspace{-2mm}
    \label{fig:special_Re-ID}
\end{figure*}

\blue{\textbf{Occluded Re-ID.} Occluded Re-ID is a variant of object Re-ID that deals with the challenge of partial occlusions in images \citep{jia2022learning,wang2022pose,xu2022learning, ye2022dynamic}. In occluded Re-ID, part of the person or object is blocked by obstacles (e.g., other people, objects, or structures), making it harder for models to extract full identity information.} Recently Transformer-based methods have made great contributions to address the occlusion challenge of Re-ID \citep{wang2023feature, mao2023attention, zhou2022motion, cheng2022more}. Extracting partial region representation features is also a good solution to the challenge of object occlusion in Re-ID. Part-Aware Transformer (PAT) \citep{li2021diverse} is proposed to exploit the Transformer encoder-decoder architecture to capture different discriminative body parts, where the encoder is used to obtain pixel context-aware feature maps and the decoder is used to generate part-aware masks. \blue{To address the issues of misalignment and occlusion, He\textit{et al.} \citep{he2023region} proposes using CLIP \citep{radford2021learning} to capture both discriminative and invariant regional features. A region generation module is designed to automatically search for and locate more discriminative regions.} For the widely studied person object, due to the occlusion noise or the occlusion region is similar to the target, an intuitive solution is to guide local feature learning with the help of human pose information. Wang \textit{et al.} \citep{wang2022pose} also adopt the transformer encoder-decoder structure to present a pose-guided feature disentangling method. With the keypoint information captured by the pose estimator, a set of learnable semantic views are introduced into the decoder to implicitly enhance the disentangled body part features. Similarly, assisted by motion information, Zhou \textit{et al.} \citep{zhou2022motion} utilize keypoint detection and part segmentation for Transformer encoder-decoder modeling. 
Besides, some transformer studies analyze occlusion problems from other perspectives \citep{xu2022learning, cheng2022more}. Xu \textit{et al.} \citep{xu2022learning} design a feature recovery Transformer (FRT) to recover occluded features using nearby target information. Considering the similarity between the local information of each semantic feature in k-nearest neighbors and the query, FRT filters out noise to restore the occluded query feature. Cheng \textit{et al.} \citep{cheng2022more} utilize knowledge learned from different source datasets to generate reliable semantic clues to alleviate domain differences between off-the-shelf semantic models and Re-ID data. Transformer allows human parsing results to be embedded as learnable tokens into the input, where a weighted sum operation is employed to integrate parsed information from multiple sources. \blue{Besides, given that the self-attention in Transformers primarily emphasizes low-level feature correlations, it inherently limits higher-order relations among different body parts or regions, which are particularly crucial for occluded person Re-ID. To address this, Li \textit{et al.} \citep{li2024occlusion} introduces a second-order attention module, which extracts contextual information from attention weights using spectral clustering techniques.}

\textbf{Cloth-changing Re-ID.} It is a challenging Re-ID task in long-term scenarios where a person may change clothes in an unknown pattern.
Cloth-changing Re-ID is a challenge unique to persons, which is a difficult but more practical problem \citep{liu2023dual}. In this scenario, the discriminative feature representation dominated by the visual appearance of clothing will be invalid. Existing research tackling this more intricate challenge has initiated initial investigations into the application of Transformers. Lee \textit{et al.} \citep{lee2022attribute} evaluate different backbones in the cloth-changing Re-ID scenario, and Transformer demonstrated notable performance advantages when compared to CNNs. On this basis, in order to further eliminate the influence of characteristics related to clothing or accessories, an attribute de-biasing module is designed. The core idea is to use the generated attribute labels for person instances as auxiliary information and adopt a gradient reversal mechanism based on adversarial learning to learn attribute-agnostic representations.

\textbf{Human-centric Tasks.} 
The success of the general large model built by Transformer lies in its ability to handle multiple tasks. Recent work has attempted to focus on human-centric general model design to facilitate the Re-ID task.
Human-centric perception integrates visual tasks such as pedestrian detection, pose estimation, attribute recognition, and human parsing. Person Re-ID is one of the human-centric tasks. These tasks all have in common that they rely on the basic structure of the human body and the properties of body parts. Previous studies have experimentally verified that training human-centric tasks together can benefit each other \citep{ci2023unihcp}. It is challenging to unify large-scale multiple tasks into a scalable model due to the different structure and granularity of annotations and expected outputs of different tasks requiring separate output headers for each task. UniHCP \citep{ci2023unihcp} presents a flexible Transformer encoder-decoder structure to avoid task-specific output heads. The core idea is to define task-specific queries in the decoder and design a task-guided interpreter to interpret each query token independently. Outputs of the same modality share the same output unit, enabling maximum parameter sharing among all tasks while learning human-centric knowledge at different granularities. Additionally, Tang \textit{et al.} \citep{tang2023humanbench} established a large-scale dataset HumanBench, specifically designed for human-centric pre-training. To address task conflicts arising from diverse annotations in supervised pre-training, a projector-assisted hierarchical pre-training method is proposed. The core idea involves constructing a hierarchical structure: sharing the weights of the backbone across all datasets, while restricting the weights of the projector to be shared only among datasets of the same tasks, and the weights of the head to be shared for a single dataset.

On the other hand, unlabeled person data is plentiful, and researchers use Transformer's strong scalability for large-scale self-supervised training to learn human-centric representations. Considering methods such as contrastive learning or masked image modeling that failed to explicitly learn semantic information, SOLIDER \citep{chen2023beyond} uses Transformer to generate pseudo semantic labels for every token based on prior knowledge of human images and introduces a token-level semantic classification pretext task to learn a stronger human semantic representation. With the mutual promotion of large-scale learning in multiple human-centric tasks, SOLIDER can achieve superior performance compared with the state-of-the-art unsupervised pre-training methods \citep{luo2021self, zhu2022pass} for Re-ID, which can be regarded as a further advance in the Re-ID community.

\textbf{Person Search.}
Person search is an end-to-end method that aims to jointly solve the two sub-problems of detection and person Re-ID using more efficient multi-task learning methods. Since the goals of person detection and person Re-ID are conflicting and it is difficult to jointly learn a unified feature representation, Yu \textit{et al.} \citep{yu2022cascade} propose to decompose feature learning into successive steps in the T stage of a multi-scale Transformer to gradually learn from coarse to fine embedding. Unlike some existing multi-scale Transformers that learn different scale information based on patches of different sizes, they leverage a series of convolutional layers with different kernels to generate multi-scale tokens. Furthermore, to produce more occlusion-robust representations, they design to exchange partial tokens of instances in mini-batches, and then compute occlusion attention based on mixed tokens. Different from the shuffling and regrouping strategy in TransReID \citep{he2021transreid}, they are for partial tokens in a single instance. PSTR \citep{cao2022pstr} is also designed as a person search multi-scale learning scheme of the Transformer architecture. It develops a PSS module consisting of a detection encoder-decoder and a discriminative re-identification decoder, where the detection encoder-decoder employs backbone features and three cascaded decoders are employed. The Re-ID decoder takes a feature query from one of the three detection decoders as input, and a multi-level supervision scheme is designed to provide different input Re-ID feature queries and box sampling locations. In order to achieve multi-scale expansion, the features of different layers use PSS modules and are concatenated to perform instance-level matching with queries.

\textbf{Group Re-ID.} It is a group-level Re-ID by using the contextual information to match a small number of individuals in a group together \citep{bmvc09group}. 
Since people usually have group and social attributes, group actions are preferred in most real-world scenarios. Group Re-ID has gradually attracted the attention of researchers, which needs to deal with challenges such as membership and layout changes. Existing group Re-ID methods are mainly based on the combined framework of CNN and GNN. However, these structures are deficient in position modeling and have weak ability to describe group layout characteristics. Inspired by the position embedding in the transformer, Zhang \textit{et al.} \citep{zhang2022uncertainty} design the second-order Transformer model SOT to deal with the layout features in group Re-ID. It consists of intra-member and inter-member modules, where each member in the group image is first cropped, and then each member is segmented into multiple sub-patches. The intra-member module extracts first-order labels as per-member features by modeling the relationship between sub-patches through a transformer. The member-to-member module models the relationship between members through uncertainty and extracts second-order tokens through transformers.

\begin{figure}
    \centering
    \includegraphics[width=\linewidth]{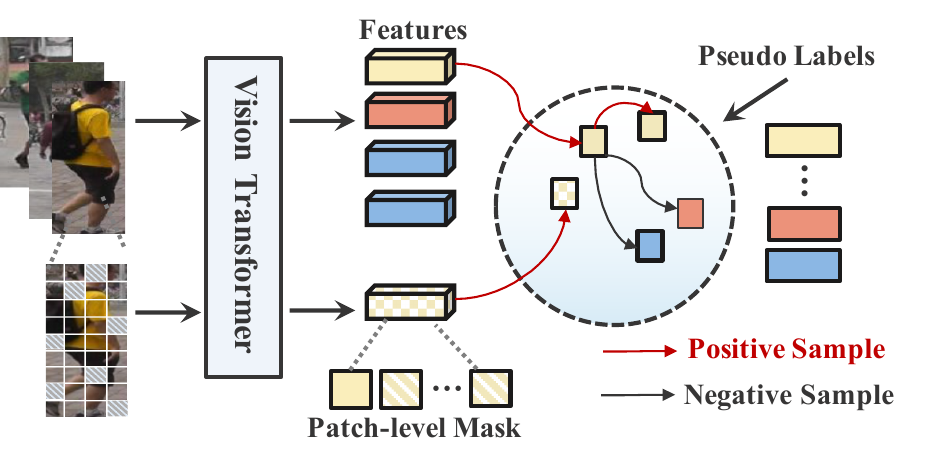}
     \vspace{-3mm}
    \caption{The proposed unsupervised Transformer baseline for Re-ID enhanced with a patch-level mask learning strategy.}
    \label{fig:usl-method}
\end{figure}

\textbf{Re-ID in UAVs.} Object Re-ID in UAVs involves identifying specific objects within a multitude of aerial images captured from a dynamic bird's-eye view \citep{organisciak2021uav}. It also has broad application prospects in different scenarios.
Re-ID using aerial images captured by UAVs is an under-explored scenario. Unlike the widely used fixed cameras, the images captured by UAVs are more complex than fixed city cameras. Unavoidable continuous rapid movement and height changes lead to large differences in image viewing angles. Chen \textit{et al.} \citep{chen2022rotation} analyzed the vehicles and persons in the bird's-eye view and concluded that Re-ID faces two key challenges in UAV scenarios: bounding boxes with significant size differences and objects with uncertain rotation direction. Benefiting from the corresponding relationship between image patches and token-level features in Vision Transformer, the insight of this work is to simulate rotation operations on the initially learned patch features to generate enhanced diversity rotation features. Another study \citep{ferdous2022uncertainty} explores the enhancement of the Pyramid Vision Transformer, leveraging multi-scale features for object Re-ID in UAV scenarios. \blue{In addition to fully drone-view Re-ID, some research focuses on the cross-view matching problem between aerial and ground perspectives. \citep{zhang2024view} proposed a view-decoupled transformer to specifically address the significant view discrepancy, aiming to decouple view-related and view-independent components.}


\section{New Unsupervised Transformer Baseline }
\label{sec:new_baseline}
After conducting a thorough review of Transformer's work in Re-ID, we are confident that large-scale pre-trained Transformers hold substantial promise for unsupervised Re-ID and warrant further exploration. Most previous Re-ID works are pre-trained on ImageNet, due to the lack of large-scale person datasets. In fact, pre-training on person datasets, such as LUPerson \citep{fu2021unsupervised}, is better suited for Re-ID task and aligns with future development trends. Our survey reveals that some studies \citep{luo2021self, zhu2022pass} verify the evident advantages of using Transformer pre-training on LUPerson. In order to further promote the progress of the Re-ID community, we propose a single/multi-modal general unsupervised Re-ID baseline. Specifically, our baseline follows the Re-ID method \citep{dai2022cluster} of contrastive learning of pseudo-labels generated by clustering, and uses the TransReID-SSL \citep{luo2021self} pre-trained Transformer as a powerful initialization model.
On this basis, leveraging the characteristics of Transformer, we have devised the following design for UntransReID.

\textbf{Single-modal Unsupervised Re-ID.}
Inspired by existing Transformer-based masked image modeling self-supervised methods \citep{he2022masked, xie2022simmim}, we design a patch-level mask enhancement strategy integrated into the unsupervised training process. Our core idea is to adopt a series of learnable tokens to mask part of the image patches as an augmentation and establish the relationship between the original features and the mask features during the training process as a supervisory signal to guide model learning. On the other hand, aligning the mask features with the original features inherently encourages the model to learn local fine-grained information. 
For input images, we define the set $\mathcal{X}^g = \{x^g_{i}|i=1,2,\cdots,n\}$ and set $\mathcal{X}^l = \{x^l_{i}|i=1,2,\cdots,n\}$ respectively as the original input and mask enhancement input. Patch embedding operations are used to get preliminary tokens $x^g_{i} \in \mathbb{R}^{N \times D}$ and $x^l_{i} \in \mathbb{R}^{N \times D}$, where N and D represent the number of patches and the dimension of the token. We initialize a set of learnable mask tokens $\mathcal{M}^l = \{m^l_{i}|i=1,2,\cdots,m\}$, randomly replacing $p$ of the tokens in $\mathcal{X}^l$ as the final input. The corresponding output class tokens after Transformer model learning are $\{f^g_{i}|i=1,2,\cdots,n\}$ and $\{f^l_{i}|i=1,2,\cdots,n\}$. We calculate the contrastive loss between the original image features and mask-enhanced features as:
\vspace{-1mm}
\begin{equation}
{\mathcal{L} = -\log\frac{\exp(f^g_{i} \cdot f^l_{i}/\tau)}{\sum_{j=1}^k \exp(f^g_{i}\cdot f_j/\tau)}}, 
\end{equation}
where $k$ represents the batch size and $f_j$ represents the original features within the batch.


\begin{table}[t]
\scriptsize
\renewcommand{\arraystretch}{1.5}
    \centering
    \caption{\label{tab:uvi-baseline} Evaluation results of our Transformer-based visible-infrared cross-modal unsupervised Re-ID baseline on RegDB \citep{nguyen2017person} and SYSU-MM01 \citep{wu2017rgb}. }
     \vspace{-3mm}
    \resizebox{0.5\textwidth}{!}{
    \begin{tabular}
    {c|cccc|cccc}
    \hline
     & \multicolumn{4}{c}{RegDB} & \multicolumn{4}{c}{SYSU-MM01} \\ \hline
    Method & \multicolumn{2}{c}{V-T} &\multicolumn{2}{c}{T-V}  & \multicolumn{2}{c}{All Search} & \multicolumn{2}{c}{Indoor Search} \\ \hline
        ~ & mAP & R1  & mAP & R1  & mAP & R1& mAP & R1\\ \hline
        OTLA  & 29.7 & 32.9 & 28.6 & 32.1 & 27.1 & 29.9& 38.8 & 29.8 \\ 
        ADCA  & 64.1 & 67.2 & 63.8 & 68.5 & 42.7 & 45.5 & 59.1 & 50.6 \\
        ACCL  & 65.4 &69.5 & 65.2 & 69.9  & 51.8 & \textbf{57.3} & 62.7 & 56.2 \\ 
        \hline
        UntransReID & \textbf{69.9} & \textbf{76.3} & \textbf{69.3} & \textbf{76.8} & \textbf{52.5} & 51.9 & \textbf{66.0} & \textbf{57.5}\\ \hline
    \end{tabular}
    }
\end{table}

\textbf{Cross-modal Unsupervised Re-ID.} For the transformer-based unsupervised visible-infrared cross-modal Re-ID, we devise a dual-path transformer that adopts two modality-specific patch embedding layers and a modality-shared transformer. Each modality-specific patch embedding layer comprises an IBN-based Convolution Stem (ICS) \citep{luo2021self} to capture modality-specific information. The modality-shared transformer is introduced to learn a multi-modality sharable space. On the basis of \citep{dai2022cluster}, two modality-specific memories are constructed for mining inter- and intra-class information within each modality with contrastive learning. To further ensure the modality generalization capability, we adopt random channel augmentation following \citep{ye2021channel} as an extra input to the visible stream for joint learning.

\textbf{Analysis of Results.}
Table \ref{tab:usl-performance} and Table \ref{tab:uvi-baseline} respectively present the evaluation results for our single-modal and cross-modal unsupervised Re-ID baselines. For single-modal unsupervised Re-ID, the structural characteristics of the Transformer allow us to generate augmented samples by applying local masks at the patch level, enabling the construction of supervisory signals. On the powerful Transformer backbone pretrained on LUPerson \citep{luo2021self}, our baseline combined with the enhancement strategy and contrastive learning \citep{dai2022cluster}, achieves performance comparable to state-of-the-art methods. For cross-modal Re-ID, three methods OTLA \citep{wang2022optimal}, ADCA \citep{yang2022augmented}, and ACCL \citep{wu2023unsupervised}, are compared.
Existing state-of-the-art methods are based on CNNs and require complex cross-modal association designs, whereas our Transformer baseline achieves state-of-the-art performance with a simple design across several infrared-visible Re-ID datasets.

\begin{figure*}[t]
    \centering
    \includegraphics[width=\linewidth]{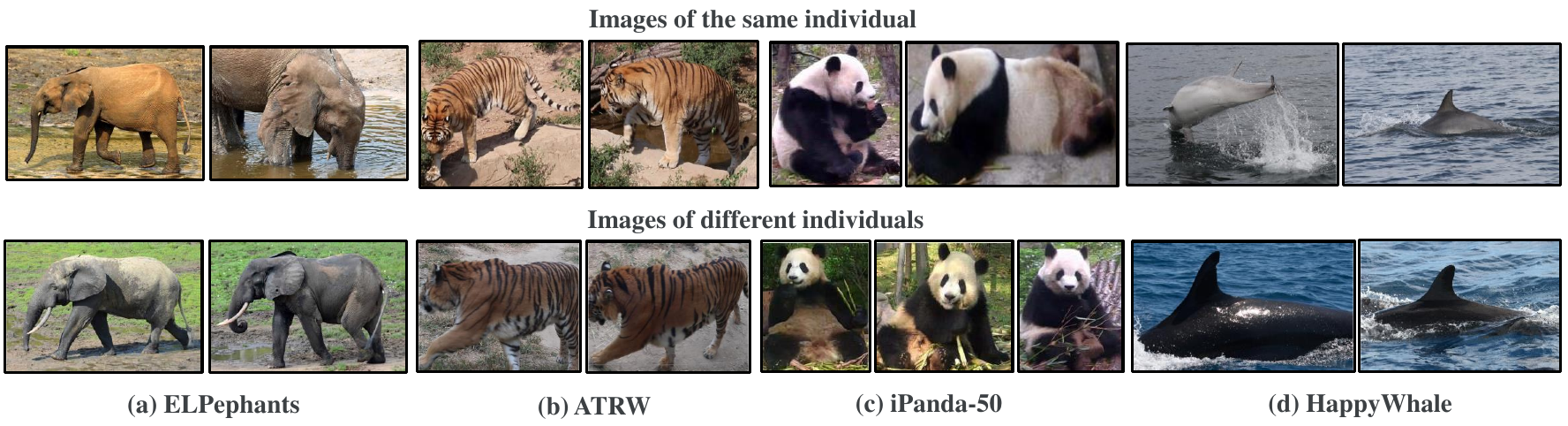}
     \vspace{-5mm}
    \caption{\textbf{Images of different species in Animal Re-ID.} Unlike the widely studied person and vehicle Re-ID, animal individuals of the same species have extremely similar appearances. Different species have their own unique discriminative characteristics, such as (a) the tusks and injury marks of elephants \citep{korschens2019elpephants}, (b) the coat pattern of amur tiger\citep{li2019atrw}, (c) the eyes of giant pandas \citep{wang2021giant}, (d) the dorsal fin, back, and flank of whales \citep{happy-whale-and-dolphin}.}
    \label{fig:animal}
\end{figure*}

\section{Animal Re-Identification}
\label{sec:animal_Re-ID}

In addition to objects such as persons and vehicles, which are currently widely studied in the field of Re-ID, the demand for wildlife protection makes animal re-identification gradually attract the attention of researchers \citep{kuncheva2022benchmark, schneider2019past}. Animal Re-ID has important applications in many fields such as ecology, conservation biology, environmental monitoring, popular science education, and agriculture. It provides a powerful tool for understanding nature, protecting biodiversity, and maintaining ecological balance. Considering that the existing Re-ID surveys are all for persons or vehicles, our survey will cover a wider range of Re-ID objects including diverse animals to promote Re-ID research. This section provides an overview of recent work concerning animal Re-ID.

Different from the mature development of person and vehicle re-identification technology, animal re-identification technology is still in a relatively early stage. One of the most intuitive challenges is that the uncontrollable environmental factors in the wild make it very complicated to collect animal images and label their individual information. 
Compared to humans, animal species exhibit a diverse array of unique characteristics across different species.
In the supplementary, we provide a more visually intuitive comparison of Re-ID across different species. The core problem of animal individual re-identification is to mine and analyze the discriminative features specific to a species.
Our survey summarizes the animal Re-ID datasets released in recent years in \S \ref{sec:animal_datasets}. In addition, deep learning-based animal re-identification techniques are introduced in \S \ref{sec:animal_methods}.


\subsection{Animal Re-ID Datasets}
\label{sec:animal_datasets}
Due to the diversity of environments and ways in which different animals live, the collection of animal data is not as simple as that of persons and vehicles. In addition to using surveillance cameras and ordinary digital cameras, camera traps, drones, and infrared thermography are also important devices. Therefore, most animal Re-ID datasets focus on annotating identity information in order to complete specific individual recognition, without a clear definition of camera. The emergence of more and more animal Re-ID datasets in recent years has promoted research progress \citep{li2019atrw, zhang2021yakReID, nepovinnykh2022sealid}. As shown in Table \ref{tab:animal}, we provide an overview of animal Re-ID datasets in recent years. We present key features of different species in Re-ID, which are also special challenges in the animal Re-ID task.
In addition, some animal data such as long-lived species are collected over a period of several years. This kind of data with a long time span provides more comprehensive information and supports deeper analysis for long-term ecological research and species protection \citep{papafitsoros2022seaturtleid}. This implies that Re-ID will be more challenging since the appearance of animals can change dramatically over time. We also report the time span for each dataset.

\begin{table*}[t]
\renewcommand{\arraystretch}{1.5}
    \centering
    \caption{\label{tab:animal} \textbf{Summary of animal Re-ID datasets from recent years.}}
     \vspace{-3mm}
    \resizebox{\textwidth}{!}{
    \begin{tabular}
    {c|ccccccc}
    \hline
    Dataset & Species & IDs & Images & Feature & Span & Source & Available \\ 
    \hline
    CattleRe-ID\citep{bergamini2018multi} & cattle & - & -  & face & - & farm & \ding{53}\\ 
    DolphinRe-ID\citep{bouma2018individual} & dolphin & 185 & 3544  & fin & 12 years & - & \ding{53}\\
    Elpephants\citep{korschens2019elpephants} & elephant & 276 & 2078  & body, tusk & 15 years & national park & \ding{53}\\ 
    ATRW\citep{li2019atrw} & Amur tiger & 92 & 3649 & stripe & - & wild zoos  & \ding{51}\\ 
    zebrafishRe-ID\citep{bruslund2020re} & zebrafish & 6 & 2224 & side view & - & laboratory  & \ding{51}\\
    CowRe-ID\citep{li2021individual} & cow & 13 & 3772  & coat pattern & -  & farm  & \ding{51}\\
    YakRe-ID-103\citep{zhang2021yakReID} & yak & 103 & 2247  & horn & - & highland pastures  & \ding{53}\\
    Cows2021 \citep{gao2021towards} & cattle & 182 & 13784  & coat pattern & 1 month  & farm  & \ding{51}\\
    iPanda-50 \citep{wang2021giant} & giant panda & 50 & 6874  & local & - & Panda Channel & \ding{51}\\
    SealID\citep{nepovinnykh2022sealid} & seal & 57 & 2080  & pelage pattern & 10 years & Lake Saimaa  & \ding{51}\\
     FiveVideos \citep{kuncheva2022benchmark} & pigeon,fish,pig & 93 & 20490  & - & - & Pixabay  & \ding{51}\\
    BelugaID \citep{belugaid2022} & beluga whale & 788 & 5902  & scarring pattern & 4 years & Cook Inlet & \ding{51}\\
    Honeybee \citep{chan2022honeybee} &  honeybee & 181 & 8962  &  abdomen & multiple weeks & colony entrance & \ding{53}\\
    HappyWhale \citep{happy-whale-and-dolphin} &  30 species & 15587 & 51033 & fin,head,flank & very long & 28 organizations & \ding{51}\\
    SeaTurtleID \citep{papafitsoros2022seaturtleid} &  sea turtle & 400 & 7774 & - & 12 years & Laganas Bay & \ding{51}\\
    LeopardID\citep{leopard-id-2022} & African leopard  & 430 & 6795 & spot pattern & 11 years & - & \ding{51}\\
    HyenaID \citep{hyena-id-2022} & spotted hyena  & 256 & 3104  & spot pattern & - & - & \ding{51}\\
    PolarBearVidID 
    \citep{zuerl2023polarbearvidid}& polar bear & 13 & 138,363  & - & - & zoo & \ding{51} \\
    Wildlife-71 \citep{jiao2023toward}& 71 species & $\approx$2059 & $\approx$108,808  & - & - & internet & \ding{51} \\
    \hline        
    \end{tabular}}
\end{table*}

Our survey collects animal Re-ID datasets from different sources in recent years to promote Re-ID research, some of which are paper publications \citep{korschens2019elpephants, nepovinnykh2022sealid, zuerl2023polarbearvidid, wang2021giant, zhang2021yakReID}, and some in the form of competitions \citep{happy-whale-and-dolphin,  li2019atrw}. This is mainly due to the fact that the datasets of many papers are not publicly available, and datasets from diverse sources allow advanced research to be conducted on them. First, some domestic or laboratory animal datasets are introduced. 
Bergamini \textit{et al.} \citep{bergamini2018multi} collect cattle head images in farms for Re-ID, considering that the cattle heads can show sufficient texture, shape and patch characteristics. Li \textit{et al.} \citep{li2021individual} shot at a real cattle farm and built a dataset of 13 cows. The Cows2021 dataset \citep{gao2021towards} contains 186 Holstein-Friesian cattle, which took a month to capture from a bird's-eye view on the farm. For cattle, their personalized black and white coat pattern patches are an important distinguishing characteristic of individuals. YakRe-ID-103 \citep{zhang2021yakReID} is a Yak dataset of highland pasture scenes. Yaks mostly have black fur and are typically texture-less animals, making it difficult to distinguish individuals. The most unique features are the thickness, bending and direction of the horn. Besides, a dataset of six zebrafish recorded in a laboratory setting is presented by \citep{bruslund2020re}. They propose reliable re-identification through the stripes of zebrafish from the side view.

Re-ID for wild animals is relatively more challenging due to complex animal habits and uncontrollable environments. ATRW \citep{li2019atrw} is a dataset containing 92 Amur tigers collected in multiple large wild zoos with bounding boxes, pose key points and identity annotations. 
Korschens \textit{et al.} \citep{korschens2019elpephants} collected images of forest elephants in national parks and constructed a data set containing 2078 images of 276 elephant individuals. The dataset spans approximately 15 years, which reflects some of the aging effects and dramatic changes in the physique of elephants. 
The most identifiable tusks and scars of elephants also change over time, exacerbating the difficulty of Re-ID. Chan \textit{et al.} \citep{chan2022honeybee} constructed a short-term and long-term dataset for honeybee re-identification.  They mainly focused on the abdomen of honeybee for individual discrimination, and the time span of the long-term dataset reached 13 days. iPanda-50 \citep{wang2021giant} is a giant panda Re-ID dataset collected through giant panda streaming videos, which contains 50 giant pandas of different ages including cubs, juveniles, and adults. The Saimaa ringed seal is an endangered subspecies found only in Lake Saimaa, Finland. Individual ringed seals have unique fur patterns, and individual re-identification is of great value in monitoring endangered animals \citep{nepovinnykh2022sealid}. SealID \citep{nepovinnykh2022sealid} is a benchmark for Saimaa ringed seal re-identification, which takes into account challenges such as the deformable nature of seals and low contrast between the ring pattern. SeaTurtleID \citep{papafitsoros2022seaturtleid} is a large-scale dataset containing images of sea turtles captured in the wild. This dataset is time-stamped and spans up to 12 years. Considering the impact of timestamps for unbiased evaluation of animal Re-ID methods, the dataset also provides time-aware partitioning of reference and query sets.
Happywhale \citep{happy-whale-and-dolphin} is an open-source platform to facilitate the identification of individual marine mammals. PolarBearVidID \citep{zuerl2023polarbearvidid} provides a video-based dataset of polar bears, which is challenging because individuals lack significant unique visual features. While the majority of existing datasets are tailored to a single species, recent research has introduced large-scale Re-ID datasets that encompass multiple species. The Wildlife-71 dataset \citep{jiao2023toward}, proposed as a dataset that aggregates existing datasets and partial web data, includes Re-ID data from 71 different wildlife categories. Indeed, it is observed that many existing animal datasets are relatively small in scale, and aggregating multi-species data proves advantageous for deep learning technologies. In addition, we observe that the animal Re-ID is much less explored compared with other objects. Recognizing that animals suffer from additional severe occlusions and viewpoint changes compared with persons.

\subsection{Animal Re-ID Methods}
\label{sec:animal_methods}
In this survey, we mainly focus on advanced deep learning methods for animal Re-ID due to their powerful performance compared with other traditional solutions. Our survey broadly categorizes these methods into three groups: learning with global animal images, learning with key local body areas, and learning with auxiliary information. Note that Transformer is seldom explored in this area. 

\textbf{Global Image Based Methods.} 
Many existing studies draw upon the conventional approaches of person Re-ID, directly feeding entire animal images into deep neural networks to acquire reliable feature representations \citep{bouma2018individual}. Taking cues from person Re-ID methodologies like local maximal occurrence \citep{liao2015person}, Bruslund \textit{et al.} \citep{bruslund2020re} introduce two feature descriptors consisting of color and texture to reliably re-identify zebrafish from side-view. Considering that the patterns on manta rays are usually in uncertain positions, Moskvyak \textit{et al.} \citep{moskvyak2021robust} devise a loss function to minimize the distance between the same individual observed from various viewpoints, guiding the learning of pose-invariant features. Porrello \textit{et al.} \citep{porrello2020robust} propose a general view knowledge distillation method for Re-ID tasks. The core idea is to use the diversity of the target in different views as a teaching signal, allowing students to use fewer views to restore it and learn more robust features.
For giant panda re-identification, Wang \textit{et al.} \citep{wang2021giant} design a multi-stream structure to learn local and global features. In order to mine local fine-grained information from the global image, a patch detector is adopted to automatically capture the most discriminative local patches without additional part annotations. 

\textbf{Local Area Based Methods.} 
Among the related work on animal Re-ID, some research focuses on specific parts of the animal. They extract the most discriminative areas of the original image during the data collection stage, such as the head of a cow \citep{bergamini2018multi}, elephant ears \citep{weideman2020extracting}, whale tails \citep{cheeseman2022advanced}, dolphin fins \citep{bouma2018individual, weideman2017integral, konovalov2018individual}, etc.
Bergamini \textit{et al.} \citep{bergamini2018multi} employ CNNs for direct feature extraction from self-collected cattle head datasets and utilize KNN (k-nearest neighbors) for classification. For fine-grained images of elephant ears and whale tails, Weideman \textit{et al.} \citep{weideman2020extracting} designed their approach to extract boundary information from color and texture transitions, along with intensity variations, to effectively discern the outlines of critical regions.


\textbf{Auxiliary Information Based Methods.} 
Zhang \textit{et al.} \citep{zhang2021yakReID} utilize a simplified definition of the pose of the yak's right or left head as an auxiliary supervision signal to enhance feature learning. Li \textit{et al.} \citep{li2019atrw} employed the results of pose key point estimation to model the tiger image into 7 parts including the trunk, front legs, and hind legs to learn local features. In order to learn the unique body markings of animal individuals with similar appearance, Moskvyak \textit{et al.} \citep{moskvyak2020learning} proposed a heat map enhancement method to display the location information of introduced animal landmarks in the Re-ID model. When dealing with species exhibiting similar pelage or fur patterns, Nepovinnykh \textit{et al.} \citep{nepovinnykh2020siamese} employed the Sato tubeness filter to extract the fur pattern from the image, mitigating the impact of interfering factors like lighting. Siamese networks \citep{koch2015siamese} trained with triplet loss are used for subsequent matching.

\subsection{A Unified Benchmark for Animal Re-ID}

In fact, existing deep learning-based animal Re-ID methods are still in the early stages of development, and we generally summarize their main limitations:
(1) \textit{Unclear Task Boundaries.} Many animal-related studies do not have clear task definitions, some of which are regarded as fine-grained recognition or individual classification \citep{wang2021giant}. They typically concentrate solely on distinguishing between different individuals and pay little attention to whether they can be reliably re-identified across various settings. However, in this survey, we emphasize animal re-identification, with the goal of accurately identifying the same individual across different timeframes, environments, or viewpoints. 
(2) \textit{Limited method applicability.} 
Many existing methods leverage the distinctive traits of particular species to develop their approaches, with some specifically focusing on curating datasets for certain body parts of the animals \citep{weideman2017integral, 
 nepovinnykh2020siamese, bergamini2018multi, chan2022honeybee}. These approaches prove challenging to adapt for broader application in Re-ID across different species and exhibit limited scalability.
(3) \textit{Inconsistent experimental settings.} 
Existing animal Re-ID methods adopt varying experimental settings.
Some of the work is conducted experimentally in a closed-world setting, which involves identifying objects within known and limited categories. In most cases, the Re-ID system is not aware of all possible categories during training, necessitating its ability to handle unforeseen categories. Some research has also been conducted in scenarios that better align with the open-set nature of Re-ID tasks. This makes performance comparison between different methods a challenging task.



To further advance research and realize the full potential of animal re-identification for practical applications, it is critical to establish standardized benchmarks and develop more robust, scalable techniques. In this survey, we conducted extensive animal Re-ID experiments using multiple state-of-the-art general Re-ID methods to address the aforementioned issues. Our work in this section covers a unified evaluation setting, a comparison of different backbone methods, and an analysis of the Transformer's suitability for animal Re-ID. The code will be publicly available.

\subsubsection{Animal Re-ID experiments.}

\textbf{Datasets.}
We chose datasets featuring various species, such as giant pandas \citep{wang2021giant}, elephants \citep{korschens2019elpephants}, seals \citep{nepovinnykh2022sealid}, giraffes \citep{parham2017animal}, zebras \citep{parham2017animal}, leopards \citep{leopard-id-2022} and tigers\citep{li2019atrw} for our evaluation. Since the datasets only provide original images and corresponding identity annotations, we uniformly divide them into training sets and test sets for the Re-ID task. Specifically, we divide each data set into 70\% of all identities as training data, and the remaining 30\% as test data. Ensure that the identities of the test set have not appeared in the training set. In the testing phase, we regard each image in the test set as a query, and all images in the test set except the query image constitute the gallery. The results for FiveVideos in Table \ref{tab:animal-results} are obtained using only pig data. The results for GZGC-G and GZGC-Z are using giraffe data and zebra data, respectively.

\begin{table}[t]
    \centering
    \caption{\label{tab:animal-results} Evaluation results of state-of-the-art Re-ID methods on multiple animal datasets. Three methods, BoT \citep{luo2019strong}, TransReID \citep{he2021transreid} and RotTrans \citep{chen2022rotation}, are compared.}
    \resizebox{0.5\textwidth}{!}{
    \begin{tabular}{c|ccc|ccc|ccc}
    \hline
       -  &\multicolumn{3}{c}{BoT} & \multicolumn{3}{c}{TransReID} & \multicolumn{3}{c}{ RotTrans} \\ \hline
        Dataset & \scriptsize{mAP} & \scriptsize{R1} & \scriptsize{mINP} & \scriptsize{mAP} & \scriptsize{R1} & \scriptsize{mINP}  & \scriptsize{mAP} & \scriptsize{R1} & \scriptsize{mINP} \\ \hline
    \scriptsize{iPanda-50}& 28.4 & 72.5 & 9.8 & 37.9& 88.8 & 10.5 & 42.6 & 91.7 & 12.9\\
    \scriptsize{ELPephants}& 15.8 &32.3 &4.5 & 15.3 & 40.2 & 3.1 & 30.2 & 56.0 & 9.7 \\
    \scriptsize{SealID} & 49.1  & 82.2 & 7.2  & 42.6 & 82.8 & 6.3 & 48.3 & 83.5 & 7.4 \\
    \scriptsize{ATRW} & 65.2  & 98.4 & 32.5 & 64.1 & 98.3 & 33.0 & 66.9 & 97.9 & 35.4 \\
    \scriptsize{GZGC-G} & 47.4  & 46.7 & 38.4 & 49.1 & 48.9 & 39.0 & 48.9 & 47.8 & 40.4 \\
    \scriptsize{GZGC-Z} & 13.7  & 23.5 & 5.6 & 16.3 & 26.0 & 7.4 & 16.2 & 26.7 & 7.2\\
     \scriptsize{LeopardID} & 27.3  & 60.1 & 9.9 & 31.6 & 63.7 & 12.5 & 32.5 & 63.0 & 13.3 \\
    \hline
    \end{tabular}
    }
\end{table}

\textbf{Evaluation Metrics.} The performance is evaluated by two widely used metrics in Re-ID tasks: Cumulative Matching Characteristic (CMC) and the mean Average Precision (mAP). It's worth noting that in the context of person and vehicle Re-ID, correctly matched objects captured by the same camera are typically excluded from the evaluation, while only objects captured by different cameras are considered. However, in animal Re-ID, where explicit camera information is often lacking in most animal datasets, we calculate all correctly matched objects uniformly. In many cases, simple samples with small viewing angle changes will lead to high Rank-k accuracy. Therefore, we calculate the metric mean Inverse Negative Penalty (mINP) \citep{ye2021deep}, which reflects the cost of finding the hardest matching sample.

\textbf{Analysis of Results.}
To evaluate the performance of different backbones in animal Re-ID, two Re-ID methods that are generally applicable to various objects, CNN-based BoT \citep{luo2019strong} and Transformer-based TransReID \citep{he2021transreid}, are employed in our experiments. As shown in Table \ref{tab:animal-results}, the mean accuracy of existing state-of-the-art Re-ID methods applied directly to animals is generally low. This also underscores that Animal Re-ID, distinct from the widely studied Re-ID objects, poses unique challenges and requires more targeted solutions in the future. The Transformer method performs better in most cases.
In addition, considering some characteristics of animal Re-ID that are different from conventional person Re-ID such as camera view and diverse orientations, we choose a state-of-the-art Transformer-based method of object Re-ID in UAVs which is mentioned in \S \ref{sec:special_Re-ID}, RotTrans \citep{chen2022rotation}, for evaluation. We believe that images of different species (e.g., marine and terrestrial animals) will exhibit a variety of rotation angles rather than being in a standing position as persons. Consequently, RotTrans demonstrates superiority in most animal Re-ID scenarios as a method that helps to learn rotation-invariant representations. Recently, researchers have proposed the development of a Re-ID model capable of handling any unseen wildlife category \citep{jiao2023toward}. The concept is similar to the domain generalization problem in conventional Re-ID tasks, and their solution involves leveraging larger scale and more diverse data. Differently, our benchmark is designed for the general animal Re-ID task, specifically aiming at the methodological level of multi-species applicability.


\section{Conclusion and Future Prospects}
\subsection{Under-Investigated Future Prospects}
\blue{\textbf{Re-ID Meets Large Language Models.} The integration of large language models (LLMs) into Re-ID tasks has emerged as a promising trend. In this context, Re-ID benefits from the advanced capabilities of LLMs in generating or processing textual descriptions that complement image data. In fact, several preliminary studies have already explored this direction: (1) \textit{Textual Assistance.} By generating or understanding textual descriptions of visual data, LLMs provide more detailed contextual information to enhance image-based Re-ID performance \citep{li2023clip, yang2024pedestrian}. (2) \textit{Cross-modal Image-Text Re-ID.} LLMs leverage their dual strengths in both the visual and textual modalities to align visual features with natural language descriptions, creating more robust representations for improved identification \citep{jiang2023cross, he2023vgsg}. (3) \textit{Unlabeled Data Utilization.} LLMs can automatically generate captions or labels for images in large datasets, reducing the need for manual annotation and increasing the dataset size for more effective Re-ID model training \citep{tan2024harnessing, zuo2024ufinebench}. 
(4) \textit{Semantic Understanding.} LLMs enhance fine-grained semantic understanding of image regions, especially in challenging scenarios such as occlusion or low-quality data. (5) \textit{Model Generalization.} LLMs possess strong generalization capabilities, enabling them to handle unseen categories more effectively, further improving the robustness of Re-ID systems \citep{tan2024harnessing}. The advanced exploratory work related to these developments is detailed in \S \ref{sec:text-image}. Looking ahead, LLMs hold even greater potential for various Re-ID scenarios, offering opportunities to further improve the adaptability, accuracy, and scalability of Re-ID systems across different applications.} 

\textbf{Unified Large-scale Foundation Model for Re-ID.} To meet the practical application demands of Re-ID, it primarily involves the utilization of a unified large-scale model that accommodates multi-modality and multi-object scenarios. In cross-modal Re-ID, existing research often concentrates on just two specific modalities. However, the sources of query cues are highly diverse, and exploring how to integrate information from various senses or data sources to create a genuinely modality-agnostic universal Re-ID model is a matter of significance. Transformer shows great potential in this problem, owing to its flexible handling of multi-modal inputs, robust relationship modeling capabilities, and scalability for processing large-scale data. The latest research reveals that Transformer continues to make breakthroughs in constructing multi-modal large models. For instance, Meta-Transformer \citep{zhang2023meta} is proposed to understand 12 modal information and offer a borderless multi-modal fusion paradigm. \blue{In the field of Re-ID, there have been some preliminary explorations into multi-modal unified models \citep{li2024all, he2024instruct}.} Besides, unifying various tasks related to Re-ID that target the same objective is a development direction with practical significance. In our survey, it is evident that several recent studies have made breakthroughs by utilizing Transformer models to unify diverse vision tasks centered around humans \citep{tang2023humanbench, ci2023unihcp, chen2023beyond}. Furthermore, constructing a universal model for multiple objects in Re-ID poses a significant challenge. This implies that many methods rely on specific information face difficulties. Particularly in animal Re-ID, the creation of a multi-species robust model holds great importance for practical applications. 

\textbf{Efficient Transformer Deployment for Re-ID.} Our survey demonstrates that Transformers indeed exhibit powerful performance in the Re-ID field. However, due to the substantial computational support required for self-attention calculations, the associated resource consumption is relatively high. In practical applications, such as video surveillance and intelligent security, there is an increasing demand for real-time performance and lightweight deployment of Re-ID models \citep{mao2023attention}. Balancing the preservation of the Transformer's robust performance in Re-ID with the imperative to reduce computational complexity becomes a crucial direction for future research. Besides, many pre-trained large-scale general foundations have been developed in general areas and how to efficiently transfer the general knowledge to the specific Re-ID tasks is also worth studying \citep{ding2023parameter}. Considering the catastrophic forgetting problem in large-scale dynamic updated camera network, how to efficiently fine-tune the previously learned Re-ID models to downstream scenarios is another important direction to explore \citep{pu2023memorizing,gu2023color}.

\subsection{Summary}
From our survey, it is evident that in the past three years, Transformer has experienced rapid development in the Re-ID field, particularly demonstrating strong advantages in more challenging scenarios such as multi-modal and unsupervised settings. We provide an in-depth analysis of the advantages of Vision Transformer in four aspects, corresponding to four Re-ID scenarios: 
\begin{enumerate}
\item \textit{Transformer in Image/Video Based Re-ID:} At the backbone level, the Transformer entirely relies on the attention mechanism, providing it with universal modeling capabilities for global, local, and spatio-temporal relationships. This inherent capability facilitates the effortless extraction of global, fine-grained, and spatio-temporal information in regular image and video Re-ID tasks.

\item \textit{Transformer in Re-ID with Limited Data or Annotations:} The emergence of Transformer provides more possibilities for unsupervised learning. Beyond conventional discriminative learning approaches, such as contrastive learning, a broader spectrum of self-supervised paradigms (\eg masked image modeling), has gained widespread attention and exploration. Furthermore, the Transformer exhibits superior adaptability to large-scale data, facilitating extensive self-supervised pre-training of more powerful and generalized models for addressing Re-ID with limited data or annotations.

\item \textit{Transformer in Cross-modal Re-ID:} 
Transformer provides a unified architecture to effectively handle data of different modalities, especially the connection of vision and language.
The multi-head attention mechanism possesses the capability to aggregate features across various feature spaces and global contexts, and the highly adaptable encoder-decoder structure is capable of accommodating diverse types of inputs and outputs. Consequently, the Transformer is particularly well-suited for establishing inter-modal associations and facilitating the fusion of multi-modal information in cross-modal Re-ID tasks.

\item \textit{Transformer in Special Re-ID Scenarios:} Driven by the demands of practical applications, the Re-ID field has given rise to a range of specialized and challenging scenarios, such as cloth-changing Re-ID, end-to-end Re-ID, group Re-ID, Re-ID in UAVs, and human-centric tasks. The initial exploratory efforts of Transformer in tackling these intricate challenges have showcased remarkable scalability and adaptability.
\end{enumerate}

This survey predominantly encompasses transformer-based Re-ID papers, primarily focusing on widely studied objects like persons and vehicles. 
Considering the application of Transformer in single-modal/cross-modal unsupervised Re-ID, which has not been fully explored by existing research, we present a Transformer-based baseline that achieves state-of-the-art performance on multiple single-modal/cross-modal Re-ID datasets. In particular, we explore the field of animal Re-ID, an area that continues to encounter challenges and unresolved issues. We develop unified experimental standards for animal Re-ID and evaluate the feasibility of employing Transformer in this context, laying a solid foundation for future research. Additionally, we delve into the future prospects of Transformers in Re-ID, aiming to further stimulate subsequent research.

\section*{Declarations}

\textbf{Availability of data} The authors declare that publicly available datasets are utilized for evaluating object Re-ID methods. The data supporting the experiments conducted in this study can be found in the paper. All publicly available datasets about person-related data used in this study are obtained and utilized following appropriate permissions and ethical guidelines.

\noindent\textbf{Acknowledgement.} This work is supported by National Natural Science Foundation of China (62176188, 62361166629, 62225113)).

\end{sloppypar}






\bibliographystyle{spbasic}

\end{document}